\documentclass[10pt,twocolumn,letterpaper]{article}

\usepackage{cvpr}

\usepackage{tikz}
\usetikzlibrary{positioning}

\usepackage{times}
\usepackage{epsfig}
\usepackage{graphicx}
\usepackage{tabularx}
\usepackage{amsmath}
\usepackage{amssymb}
\usepackage{bm}
\usepackage{bbm}
\usepackage{makecell}
\usepackage{color}
\usepackage{cite}
\usepackage{cuted}
\usepackage{capt-of}
\usepackage{subcaption}
\usepackage{import}
\usepackage{xspace}
\usepackage{booktabs}
\usepackage{physics} % \pdv{}{} command etc
\usepackage{soul} % \hl{} command
\usepackage{xr}
\usepackage[export]{adjustbox}
\usepackage{authblk}
\usepackage[pagebackref=true,breaklinks=true,letterpaper=true,colorlinks,bookmarks=false]{hyperref}

\newcommand{\papertitle}{FroDO: From Detections to 3D  Objects}
\newcommand{\methodtitle}{FroDO\xspace}

\newcommand{\chamfer}{D_{\text{C}}}

\newcommand{\energyg}{E_\text{g}}
\newcommand{\energyt}{E}
\newcommand{\energys}{E_\text{s}}
\newcommand{\energyr}{E_\text{r}}
\newcommand{\energyp}{E_\text{p}}
\newcommand{\bx}{\mathbf{x}}
\newcommand{\by}{\mathbf{y}}
\newcommand{\bz}{\mathbf{z}}
\newcommand{\sdf}{\phi}
\newcommand{\Gs}{G_s}
\newcommand{\Gd}{G_d}
\newcommand{\redwood}{\text{Redwood-OS}\xspace}

\newcommand{\meng}[1]{#1}
\newcommand{\todo}[1]{}
\renewcommand{\etal}{\textit{et al.}}
\newcommand{\rmCAD}[1]{}
\newcommand{\cmt}[1]{\ignorespaces}

\cvprfinalcopy % *** Uncomment this line for the final submission

\def\cvprPaperID{10713} % *** Enter the CVPR Paper ID here

\ifcvprfinal\fi
\begin{document}

\title{\papertitle}
\author[1,3,*]{Kejie Li}
\author[1,2,*]{Martin R\"unz}
\author[1]{Meng Tang}
\author[1]{Lingni Ma}
\author[1]{Chen Kong}
\author[1]{Tanner Schmidt}
\author[3]{Ian Reid}
\author[2]{Lourdes Agapito}
\author[1]{Julian Straub}
\author[1]{Steven Lovegrove}
\author[1]{Richard Newcombe}

\affil[1]{Facebook Reality Labs}
\affil[2]{Department of Computer Science, University College London}
\affil[3]{School of Computer Science, The University of Adelaide}

\twocolumn[{%
    \renewcommand\twocolumn[1][]{#1}%
    \maketitle
    \centering
    \includegraphics[width=\linewidth]{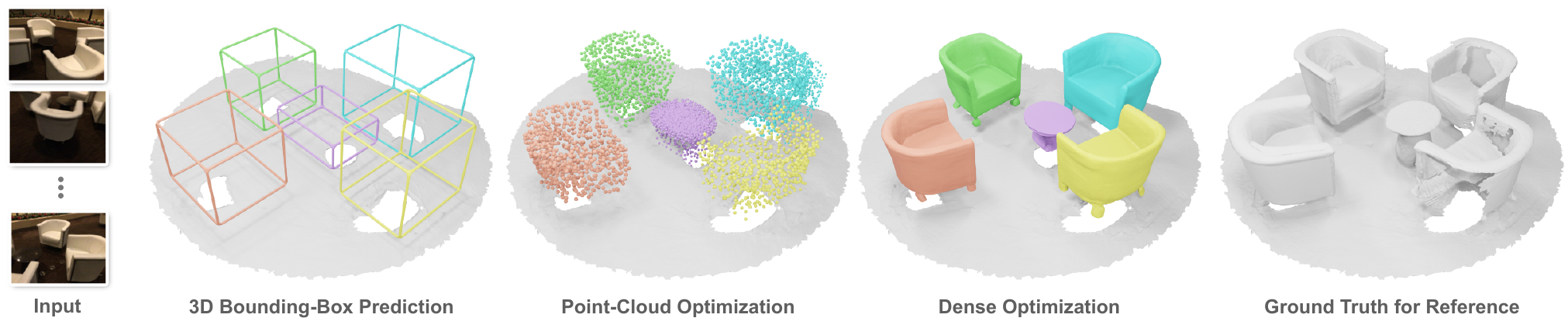}

\captionof{figure}{Given a localized input RGB sequence, \methodtitle{} dectects objects and infers their pose and a progressively fine grained and expressive object shape representation. Results on a real-world sequence from ScanNet~\cite{dai2017scannet}. %From left to right: 3D bounding-box predictions, point-cloud optimization, dense optimization and ground-truth ScanNet scan for reference.
\label{fig:teaser}}
%https://www.overleaf.com/project/5d936673619f45000176294ehttps://www.overleaf.com/project/5d936673619f45000176294e

    \vspace{1.0cm}
}]

\renewcommand*{\thefootnote}{\fnsymbol{footnote}}
\footnotetext{* The first two authors contributed equally.}
\renewcommand*{\thefootnote}{\arabic{footnote}}
\setcounter{footnote}{0}

\begin{abstract}
\vspace{-0.5em}
Object-oriented maps are important for scene understanding since they jointly capture geometry and semantics, allow individual instantiation and meaningful reasoning about objects. We introduce \methodtitle, a method for accurate 3D reconstruction of object instances from RGB video that infers object location, pose and shape in a coarse-to-fine manner. Key to \methodtitle is to embed object shapes in a novel learnt space that allows seamless switching between sparse point cloud and dense DeepSDF decoding. Given an input sequence of localized RGB frames, \methodtitle  first aggregates 2D detections to instantiate a category-aware 3D bounding box per object. A shape code is regressed using an encoder network before optimizing shape and pose further under the learnt shape priors using sparse and dense shape representations. The optimization uses multi-view geometric, photometric and silhouette losses. We evaluate on real-world datasets, including Pix3D, Redwood-OS, and ScanNet, for single-view, multi-view, and multi-object reconstruction. 
\end{abstract}

\section{Introduction}\label{sec:introduction}
Localizing and reconstructing 3D objects from RGB video is a fundamental problem in computer vision.
Traditional geometry-based multi-view reconstruction~\cite{schoenberger2016sfm,schoenberger2016mvs} can deal with large scenes given rich textures and large baselines but it is prone to failure in texture-less regions or
when the photo-consistency assumption does not hold. Besides, these methods
only provide geometry information but no semantics. An even more challenging
question is how to fill in unobserved regions of the scene.  Recently, learning based 3D 
reconstruction methods~\cite{3D-R2D2,PSGN,3D-VAE-GAN,OccupancyNetworks,gkioxari2019mesh,chabra2019stereodrnet} have
emerged and achieved promising results. 
However, data-driven approaches rely heavily on synthetic renderings and do not
generalize well to natural images. On
the other hand, we have seen impressive progress in 2D recognition tasks such as detection and segmentation \cite{he2017mask,long2015fully,kirillov2019panoptic}.

\begin{figure*}[t]
\centering
\includegraphics[width=0.95\linewidth]{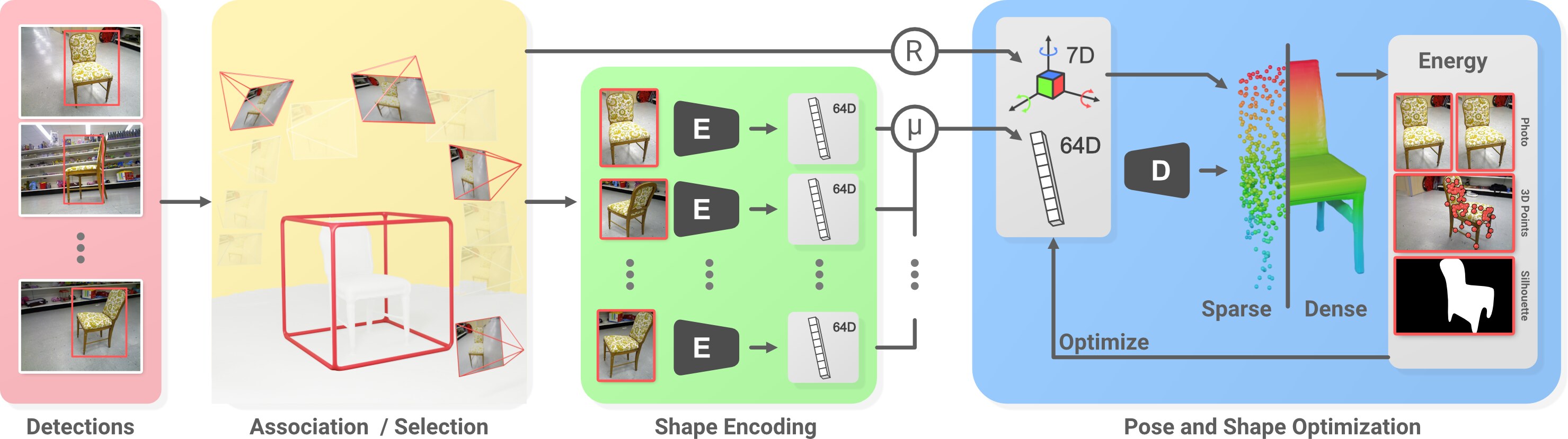}

\captionof{figure}{Given a sequence of calibrated, and localized RGB images, \methodtitle{} detects objects and infers their shape code and per-frame poses in a coarse-to-fine manner. %The steps in the process are: \emph{2D detection}$\rightarrow$ \emph{data association+view selection}$\rightarrow$ \emph{single-view shape encoding}$\rightarrow$ \emph{multi-view shape and pose optimization}.
We demonstrate \methodtitle{} on challenging sequences from real-world datasets that contain a single object (\redwood{}) or multiple objects (ScanNet).
\label{fig:pipeline}}
\end{figure*}
%https://www.overleaf.com/project/5d936673619f45000176294ehttps://www.overleaf.com/project/5d936673619f45000176294e

In this paper, we propose a system for object-centric reconstruction that leverages the best properties of 2D recognition, learning-based object reconstruction and multi-view optimization with deep shape priors. 
As illustrated in Fig.~\ref{fig:pipeline} \methodtitle
takes a sequence of localized RGB images as input, and progressively outputs 2D and 3D bounding boxes, \mbox{7-DoF} pose, a sparse point cloud and a dense mesh for 3D objects in a coarse-to-fine manner. 
\methodtitle demonstrates deep prior-based 3D reconstruction of real world multi-class and multi-object scenes from real-world RGB video. Related approaches are limited to single
view~\cite{AtlasNet,3D-VAE-GAN,PSGN,OccupancyNetworks,grabner20183d}, 
or multi-view but single objects~\cite{2019-lin}, are purely geometry-based~\cite{schoenberger2016sfm,schoenberger2016mvs}, or require depth and object scans~\cite{2019-hou-sis,2019-hou-sic}.

Choosing the best shape representation remains a key  open problem in 3D reconstruction. Signed Distance Functions (SDF) have emerged as a powerful representation for learning-based reconstruction~\cite{2019-park,OccupancyNetworks} but are not as compact or efficient as point clouds.
One of our key contributions is a new joint embedding where shape codes can be decoded to both a sparse point cloud and a dense SDF. Our joint shape embedding enables seamless switching between both representations and can be used as a shape prior for shape optimization, enabling faster inference. 
As Fig.~\ref{fig:pipeline} illustrates, \methodtitle takes a calibrated, and localized image sequence as input and proceeds in four distinct steps: \emph{2D detection}, \emph{data association}, \emph{single-view shape code inference} and \emph{multi-view shape code optimization}. First, per-frame 2D bounding box detections are inferred using an off-the-shelf method~\cite{he2017mask}. Secondly, bounding boxes are associated over multiple frames and lifted into 3D. Next, a $64$D code is predicted for each detection of the same object instance, using a novel encoder network. Per-image shape codes of the same instance are fused into a single code. Finally, shape code and pose are further refined by minimizing terms based on geometric, photometric and silhouette cues using our joint embedding as a shape prior. The final outputs of our system are dense object meshes placed in the correct position and orientation in the scene. 

The contributions of our paper are as follows:
    \emph{(i)} \methodtitle takes as input RGB sequences of real world multi-object scenes and infers an object-based map, leveraging 2D recognition, learning-based 3D reconstruction and multi-view optimization with shape priors. \emph{(ii)} We introduce a novel deep joint shape embedding that allows simultaneous decoding to sparse point cloud and continuous SDF representations, and enables faster shape optimization. \emph{(iii)} We introduce a new coarse-to-fine multi-view optimization approach that combines photometric and silhouette consistency costs with our deep shape prior.  %We introduce a new shape-prior based coarse-to-fine optimization scheme for DeepSDF decoders, enabling image-based object code and pose estimation. The reconstruction quality of our dense optimization is comparable to related approaches~\cite{liu2019dist}, while being significantly faster. 
    \emph{(iv)} \methodtitle outperforms state of the art 3D reconstruction methods on real-world datasets --- Pix3D~\cite{pix3d} for single-object single-view and  \redwood{}~\cite{2016choiredwood} for single-object multi-view. We demonstrate  multi-class and multi-object reconstruction on challenging sequences from the ScanNet dataset~\cite{dai2017scannet}.

\section{Related Work}\label{sec:related}
At its core our proposed system infers dense object shape reconstructions from RGB frames, so it relates to multiple areas in 3D scene reconstruction and
understanding.

\noindent \textbf{Single-view learning-based shape prediction}
In recent years, 3D object shape and pose estimation from images has moved from being purely
geometric towards learning techniques, which typically depend on synthetic rendering of ShapeNet~\cite{chang2015shapenet} or realistic 2d-3d datasets like Pix3d~\cite{pix3d}. These approaches can be categorized based on the shape representation utilized, for example occupancy grids~\cite{3D-R2D2,3D-VAE-GAN}, point clouds~\cite{PSGN}, meshes~\cite{wang2018pixel2mesh}, or implicit functions~\cite{OccupancyNetworks}. Gkioxari~\etal~\cite{gkioxari2019mesh} jointly train detection and reconstruction by augmenting Mask RCNN with an extra head that outputs volume and mesh.

\begin{figure}
    \centering
    \includegraphics[width=0.9\linewidth]{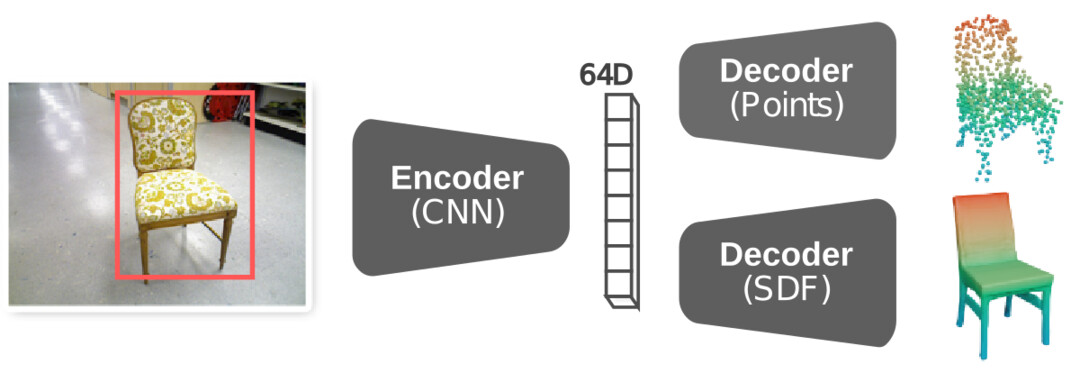}
    \caption{Our new joint shape embedding leverages the advantages of sparse point-based (efficiency) and dense surface (expressiveness)
    object shape representations.}
    \vspace{-0.5em}
    \label{fig:sharedCodeSpace}
\end{figure}

Our coarse-to-fine reconstruction pipeline includes a single-image encoder decoder network that predicts a latent shape code, point cloud, and SDF for each detected instance. Our single-view reconstruction network leverages a novel joint embedding that simultaneously outputs point cloud and SDF (Fig.~\ref{fig:sharedCodeSpace}).
Our quantitative evaluation shows that our approach provides better single view reconstruction than competing methods.

\noindent \textbf{Multi-view category-specific shape estimation}
Structure-from-Motion (SfM) and simultaneous localization and mapping (SLAM) are useful to reconstruct 3D structure from image collections or videos. 
However, traditional methods are prone to failure when there is a large gap
between viewpoints, generally have difficulty with filling featureless areas,
and cannot reconstruct occluded surfaces. 
Deep learning approaches like 3D-R2N2~\cite{3D-R2D2}, LSM~\cite{kar2017learning}, and Pix2Vox~\cite{xie2019pix2vox} have been proposed for 3D object shape reconstruction. 
These can infer object shape from either single or multiple observations using
RNN or voxel based fusion. However, these fusion techniques are slow and data
association is assumed. %Simple average in the latent space is efficient and capable of providing a good initialization for object shape in the proposed approach.

\noindent \textbf{3D reconstruction with shape priors} 
These methods are the most closely related to our approach since they also use RGB video as input and optimize object shape and pose using 3D or image-based reprojection losses such as photometric and/or silhouette terms while assuming, often category-specific, learnt compact latent shape spaces.  
Some examples of the low dimensional latent spaces used are
PCA~\cite{2019-wang,li2018optimizable}, 
GPLVM~\cite{2012-prisacariu,prisacariu2011shared,dame2013dense} or a learnt neural network~\cite{2019-lin}. 
In similar spirit we optimize a shape code for each object, using both 2D and
3D alignment losses, but we propose a new shape embedding that jointly encodes
point cloud and DeepSDF representations and show that our coarse-to-fine optimization leads to more accurate results. 
These optimizable codes have also been used to infer the overall shape of entire
scenes~\cite{2018-bloesch,sitzmann2019scene} without lifting the representation
to the level of objects.
Concurrent work~\cite{liu2019dist} proposes to optimize DeepSDF embeddings via sphere tracing, closely related to FroDO's dense optimization stage. We chose to formulate the energy via a proxy mesh, which scales better when many views are used.

\noindent \textbf{Object-aware SLAM} Although our system is not sequential or real-time, it shares common ground with recent object-oriented SLAM methods. Visual SLAM has recently evolved from
purely geometric mapping (point, surface or volumetric based) to object-level
representations which encode the scene as a collection of reconstructed object
instances.
SLAM++~\cite{salas2013slam} demonstrated one of the
first RGB-D object-based mapping systems where a set of previously known object
instances were detected and mapped using an object pose graph. Other
instance-based object-aware SLAM systems have either aligned objects from a pre-trained database to volumetric maps~\cite{Tateno2016When2I} or models learnt during an exploration step to a surfel representation~\cite{stuckler2012model}.  In contrast, others have focused on online object discovery and modeling~\cite{Choudhary2014SLAMWO} to deal with unknown object instances, dropping the need for known models and pre-trained detectors. Recent RGB-D object-aware SLAM methods leverage the power of state of the art 2D instance semantic segmentation masks~\cite{he2017mask} to obtain object-level scene graphs and per-object reconstructions~\cite{2018-mccormac} even in the case of dynamic scenes~\cite{runz2018maskfusion,xu2019mid}. Object oriented SLAM has also been extended to the case of monocular RGB-only~\cite{pillai2015monocular,nicholson2018quadricslam,parkhiya2018constructing,hosseinzadeh2019real,galvez2015realtimeMO} or visual inertial inputs~\cite{fei2018visual}. Pillai and Leonard~\cite{pillai2015monocular} aggregate multiview detections to perform SLAM-aware object recognition and semi-dense reconstruction, while~\cite{nicholson2018quadricslam} fit per-object 3D quadric surfaces.  
CubeSLAM~\cite{Yang2019CubeSLAMM3} proposes a multi-step object reconstruction pipeline where initial cuboid proposals, generated from single view detections, are further refined through multiview bundle-adjustment. 

\section{Method Overview}\label{sec:overview}
FroDO infers an object-based map of a scene, in a coarse-to-fine manner, given a localized set of  RGB images. 
We assume camera poses and a sparse point cloud have been estimated using standard SLAM or SfM methods such as ORB-SLAM~\cite{mur2015orb} or COLMAP~\cite{schoenberger2016sfm,schoenberger2016mvs}.
We represent the object-based map as a set of object poses $\{T_{wo}^k\}$ with associated 3D bounding boxes $\{bb_3^k\}$ and shape codes $\{z^k\}$. $\mathbf{T}_{ba}$ denotes a transformation from coordinate system $a$ to $b$. Our new joint shape embedding is described in Sec.~\ref{sec:joint_embedding}. 

The steps in our pipeline are illustrated in
Fig.~\ref{fig:pipeline}: First (Sec.~\ref{sec:method_detectionandassociation}) objects are detected in input images using any off-the-shelf detector~\cite{he2017mask}, correspondences are established between detections of the same object instance in different images and 2D bounding boxes are lifted into 3D, which enables occlusion reasoning for view selection. Second, a $64$D shape code is predicted for each visible cropped detection of the same object, using a novel 
convolutional neural network (Sec.~\ref{sec:method_singleviewshapeinference}). Codes are later fused into a unique object shape code (Sec.~\ref{sec:method_fusion}).
Finally, object poses and shape code are incrementally refined by minimizing energy terms based on geometric and multiview photometric consistency cues using our joint shape embedding as a prior (Sec.~\ref{sec:method_optimization}).

\section{Joint Shape Code Embedding}
\label{sec:joint_embedding}
\begin{figure*}
\centering
\includegraphics[width=\linewidth]{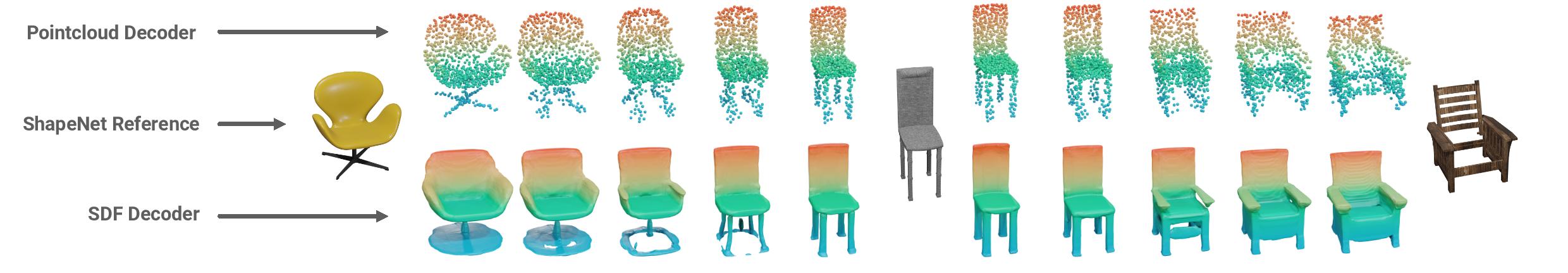}
\caption{Joint latent shape space interpolation between 3 ShapeNet instances with ground-truth codes. Pointcloud and SDF decodings of intermediate codes are coherent.
\label{fig:joint_embedding}}
\end{figure*}

We propose a new joint latent shape-code space to represent and instantiate complete object shapes in a compact way. This novel embedding is also used as a shape prior to efficiently optimize object shapes from multi-view observations.
We parametrize object shapes with a latent code $\bz  \in \mathbb{R}^{64}$, which
can be jointly decoded by two generative models $\mathbf{X} = \Gs(\bz)$ and $\sdf = \Gd(\bz)$ into 
an explicit sparse 3D pointcloud $\mathbf{X}$ and an implicit  signed distance function $\sdf$.
While the pointcloud decoder generates $2048$ samples on the object surface, the SDF decoder represents the surface densely via its zero-level set.
The decoders are trained simultaneously using a supervised reconstruction loss against ground-truth shapes on both representations:% in a similar way to DeepSDF~\cite{2019-park}: 
\begin{align}\label{eq:jointSpaceLoss}
  & L= \lambda_1 \chamfer(\Gs(\bz), \mathbf{X}_{gt}) + \lambda_2 L_{\sdf} + \frac{1}{\sigma^{2}} \|\bz\|^2, \\
  & L_{\sdf} = |clamp(\Gd(\bz), \delta) - clamp(d_{gt}, \delta)|,
\end{align}
\vspace{-1em}
\begin{align} \label{eq:distance_chamfer}
\chamfer(\mathbf{A}, \mathbf{B}) = 
     &\frac{1}{|\mathbf{A}|} \sum_{\bx \in \mathbf{A}} \min_{\by \in \mathbf{B}}\|\bx-\by\|_{2}^{2} \\
   + &\frac{1}{|\mathbf{B}|}\sum_{\mathbf{y} \in \mathbf{B}} \min_{\bx \in \mathbf{A}}\|\bx-\by\|_{2}^{2} \nonumber
\end{align}
where $\chamfer$ evaluates a symmetric Chamfer distance, and $L_{\sdf}$ is a clipped $L_1$ loss between  predicted $\Gd(\bz)$ and ground-truth $d_{gt}$ signed distance values with a threshold $\delta=0.1$. We use 3D models from the CAD model repository ShapeNet~\cite{chang2015shapenet} as ground truth shapes. While the original DeepSDF architecture~\cite{2019-park} is employed for the SDF decoder, a variant of PSGN~\cite{PSGN} is used as the pointcloud decoder. Its architecture is described in detail in the supplementary material. Joint embeddings decoded to both representations are illustrated in Fig.~\ref{fig:joint_embedding}.
The trained decoders allow us to leverage learnt object shape distributions, and act as effective priors for optimization based 3D reconstruction.
In contrast to related prior-based shape optimization approaches~\cite{2019-lin,liu2019dist} where the shape embedding is specialized to a specific
representation, our embedding offers the advantages of both sparse and dense representations at different stages of the optimization.
Although DeepSDF can represent smooth and dense object surfaces, it is slow to evaluate as each point needs a full forward pass through the decoder. 
In contrast, the pointcloud  representation is two orders of magnitude faster but fails to capture shape details.
Our strategy is therefore to infer an initial shape using the point-based decoder before switching  to the DeepSDF decoder for further refinement (Sec.~\ref{sec:method_optimization}).
While inspired by~\cite{2019-muralikrishnan} to use multiple shape representations, our embedding offers two advantages. First, the same latent code is used by both decoders, which avoids the need for a latent code consistency loss~\cite{2019-muralikrishnan}. Secondly, training a shape encoder for each representation is not required.

\section{From Detections to 3D Objects}\label{sec:method}

\subsection{Object Detection and Data Association \label{sec:method_detectionandassociation}}

We use a standard instance segmentation network~\cite{he2017mask} to detect object bounding
boxes $bb^2_i$ and object masks $\bm{M}$ in
the input RGB video. 
\begin{figure}
  \centering
  \includegraphics[width=0.4\textwidth]{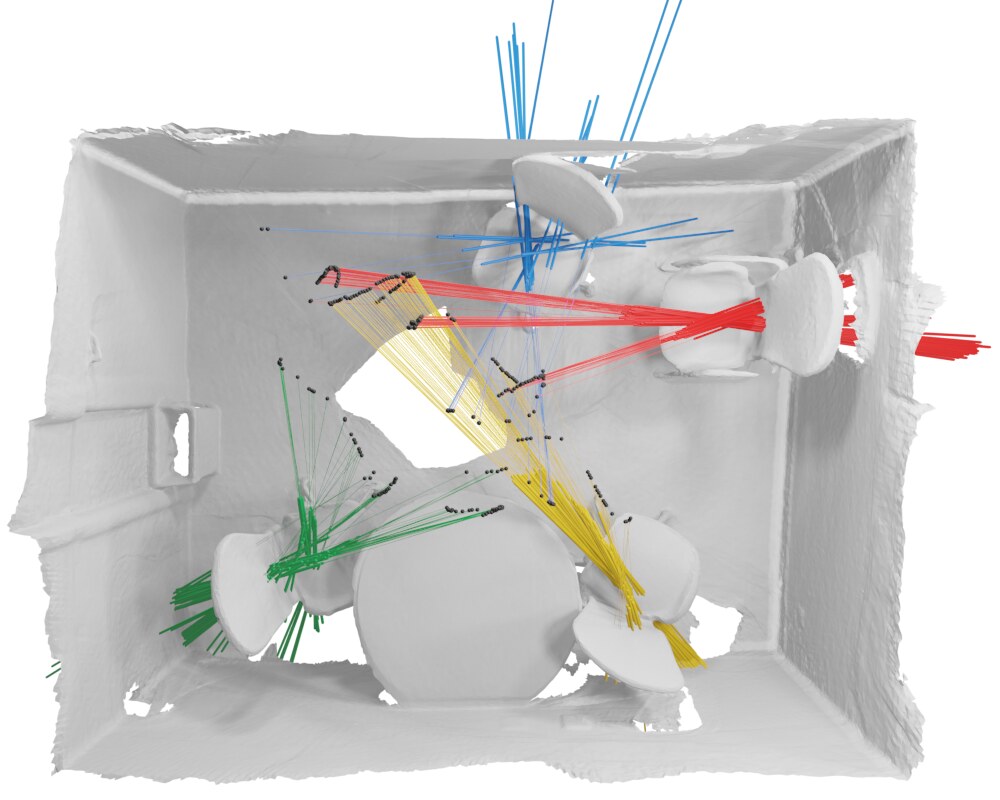}
  \caption{Data association: 3D line-segment clustering to predict b-box correspondences. Colors denote instance IDs.}\label{fig:rayCluster}
  \vspace{-0.5em}
\end{figure}
To enable multi-view fusion and data aggregation for object shape inference, we predict correspondences between multiple detections of the same 3D object instance.
Since the 3D ray through the center of a 2D bounding box points in the direction of the object center, the set of rays from all corresponding detections should approximately intersect. 
Knowledge of the object class sets reasonable bounds on
the object scale to further restrict the expected object center location in 3D to a line segment as indicated by the thicker line segments in Fig.~\ref{fig:rayCluster}.

Object instance data association can then be cast as a clustering problem in which the goal is to identify an unknown number  of line segment sets that approximately intersect in a single point.
We adopt an efficient iterative non-parametric clustering approach
similar to DP-means~\cite{kulis2011revisiting} where the observations are line segments and the cluster centers are 3D points. Further details of the clustering algorithm are given in the supplementary material. 

After clustering, each object instance $k$ is associated with a set of 2D image detections $I_k$ and a 3D bounding box $bb^3_k$, computed from the associated bounding box
detections as described in~\cite{nicholson2018quadricslam}.
By comparing the projection of the 3D object bounding box and the 2D detection box, we reject detections that have low IoU, an indication of occlusions or truncations. The filtered set of image detections $I'_k$ is used in all following steps. Examples of the filtered detections are shown in supplementary material.

\subsection{Single-view Shape Code Inference and Fusion \label{sec:method_singleviewshapeinference}}

As illustrated in the \emph{shape encoding} section of Fig.~\ref{fig:pipeline}, a $64$D object shape code is predicted for each filtered detection. We train a new encoder network that takes as input a single image crop and regresses its associated shape code $\mathbf{z_i} \in \mathbb{R}^{64}$ in the joint latent space described in Sec.~\ref{sec:joint_embedding}. 

The network is trained in a fully supervised way. However, due to the lack of 3D shape annotations for real world image datasets, we train the image encoder using synthetic ShapeNet~\cite{chang2015shapenet} renderings. Specifically, we generate training data by rendering ShapeNet CAD models with random viewpoints, materials, environment mapping, and background. We also perturb bounding boxes of rendered objects and feed perturbed crops to the encoder during training. We chose a standard ResNet architecture, modifying its output to the size of the embedding vector. During training, we minimize the Huber loss between predicted and target embeddings, which we know for all CAD models. 
\meng{For the experiment on ScanNet in Sec.~\ref{sec:experiment_scannet}, we fine-tune the encoder network with supervised data from Pix3D\cite{pix3d}.}

\noindent\textbf{Multi-view Shape Code Fusion}\label{sec:method_fusion}
For each object instance $k$ we fuse all single-view shape codes $\{\mathbf{z_i} | i\in I'_k\}$ into
a unique code $z_k^0$. We propose two fusion approaches and evaluate them in  Table~\ref{tab:ab_upgrade}: \emph{(i)} Average -- we average  shape codes to form a mean code $z_k^\text{mean}$; \emph{(ii)} Majority voting -- We find the 4 nearest neighbors of each predicted code $\mathbf{z_i}$ among the models in the training set. The most frequent of these is chosen as  $z_k^\text{vote}$. Unlike the  average code, $z_k^\text{vote}$ guarantees valid shapes from the object database.

\subsection{Multi-view  Optimization with Shape Priors\label{sec:method_optimization}}

For each object instance $k$, all images with non-occluded detections are used as input to an energy optimization approach to estimate object pose $T_{wo}^k$ and shape code $z_k$ in two steps.
First, we optimize the energy over a sparse set of surface points, using the point decoder $\Gs(\bz)$ as a shape prior. This step is fast and efficient due to the sparse nature of the representation as well as the light weight of the pointcloud decoder. Second, we further refine pose and shape minimizing the same energy over dense surface points, now using the DeepSDF decoder $\Gd(\bz)$ as the prior. This slower process is more accurate since the loss is  evaluated over all surface points, and not sparse samples.

\noindent\textbf{Energy.}
Our energy is a combination of losses on the 
2D silhouette $E_s$, photometric consistency $E_p$ and geometry $E_g$ with
a shape code regularizer $E_r$:
\begin{equation} \label{eq:total_energy}
\energyt = \lambda_s \cdot \energys + \lambda_p \cdot \energyp + \lambda_g \cdot
  \energyg + \lambda_r \cdot \energyr \,,
\end{equation}
where $\lambda_{s,p,g,r}$ weigh the contributions of individual terms.
The regularization term
$\energyr=\frac{1}{\sigma^{2}}\|\mathbf{z}\|_{2}^{2}$ encourages shape codes to take values in valid regions of the embedding, analogously to the regularizer in Eq.~\ref{eq:jointSpaceLoss}. Note that the same energy terms are used for sparse and dense optimization -- the main differences being the number of points over which the loss is evaluated, and the decoder $\mathcal{G}(\bz)$ used as shape prior. 

\noindent\textbf{Initialization.} The $64$D shape code is initialized to the fused shape code (Sec.~\ref{sec:method_fusion}), while the pose $\mathbf{T}_{wo}$ is initialized from the 3D bounding box $bb^3_k$ (Sec.~\ref{sec:method_detectionandassociation}): translation is set to the vector joining the origin of the world coordinate frame with the 3D  bounding box centroid, scale to the 3D bounding box height and rotation is initialised using exhaustive search for the best rotation about the gravity direction -- under the assumption that objects are supported by a ground-plane perpendicular to gravity. %The cost used for the  search is defined below.

\noindent\textbf{Sparse Optimization.} 
Throughout the sparse optimization, the energy $\energyt$ is defined over the sparse set of $2048$ surface points $\mathbf{X}$, decoded with the point-based decoder $\Gs(\bz)$. The energy $\energyt$ is minimized using the Adam optimizer~\cite{kingma2014adam} with autodiff. We now define the energy terms.

\noindent$\bullet$ The photometric loss $\energyp$ encourages the colour of 3D points to be consistent across views. 
In the sparse case, we evaluate $\energyp$ by projecting points in $\mathbf{X}$ to $N$ nearby  frames via known camera poses $\mathbf{T}_{cw}$ and comparing colors in reference $\mathcal{I}^{R}$ and source $\mathcal{I}^{S}_i$ images under a Huber norm~$\|.\|_{h}$:
\begin{equation} \label{eq:distance_photometric}
\begin{aligned}
&\energyp(\mathbf{X}, \mathcal{I}^R, \mathcal{I}^S_{1}, ... ,\mathcal{I}^S_{N}) = \frac{1}{N \cdot |\mathbf{X}|} \sum_{i =1}^{N} \sum_{\bx \in \mathbf{X}} \| r(\mathcal{I}^R, \mathcal{I}^S_{i}) \|_{h} \, \\
&r(\mathcal{I}^R, \mathcal{I}^S) = \mathcal{I}^{R}(\pi(\mathbf{T}_{cw}^R \bx)) - \mathcal{I}^{S}(\pi(\mathbf{T}_{cw}^S \bx))
\end{aligned}
\end{equation}
where $\pi(\bx)$ projects 3D point $\bx$ into the image. 

\noindent$\bullet$ The silhouette loss $\energys$ penalizes discrepancies between the 2D silhouette obtained via projection of the current 3D object shape estimate and the mask predicted with MaskRCNN~\cite{he2017mask}. 
In practice, we penalize points that project outside the predicted mask using
the 2D Chamfer distance:
\begin{equation} \label{eq:sparse_silhouette}
  E_s(\bz_k, \mathbf{T}_{wo}^k) = %\sum_{\mathbf{M}}%{i\in I'_k}
  \chamfer^{}(\mathbf{M}, %^i,
  \pi(\mathbf{T}_{cw} %^i
  \mathbf{T}_{wo}^k  \mathcal{G}(\mathbf{z}))) \,
\end{equation}
where $\mathbf{M}$ is the set of 2D samples on the predicted mask and $D_C$ is the symmetric Chamfer distance defined in Eq.~\ref{eq:distance_chamfer}.

\noindent$\bullet$ The geometric loss $\energyg$ minimizes the 3D Chamfer distance between 3D SLAM (or SfM) points and points on the current object shape estimate:

\begin{equation} \label{eq:sparse_geometric}
\energyg(\mathbf{z_k}, \mathbf{T}_{wo}^k) = \chamfer^{}(\mathbf{X}_{slam}, \mathbf{T}_{wo}^k \mathcal{G}(\bz)),
\end{equation}

\noindent\textbf{Dense Optimization.} 
The shape code and pose estimated with the sparse optimization can be further refined with a dense optimization over all surface points and using the DeepSDF decoder $\Gd(\bz)$. Since $\Gd(\bz)$ uses an implicit representation of the object surface, we compute a proxy mesh at each iteration, and formulate the energy over its vertices. This strategy proved faster than sphere tracing~\cite{liu2019dist}, while achieving on-par accuracy, see Table~\ref{tab:shape360}. Relevant Jacobians are derived analytically and are given in the supplementary material together with further implementation details. We now describe the dense energy terms.

\noindent$\bullet$ The photometric and geometric losses $\energyp, \energyg$ are equivalent to those used in the sparse optimization (see Eq.~\ref{eq:distance_photometric},~\ref{eq:sparse_geometric}).  However, 
they are evaluated densely and the photometric optimization makes use of a Lucas-Kanade style warp.

\noindent$\bullet$ The silhouette loss $\energys$ takes a different form to the sparse case. We follow traditional level set approaches, comparing the projections of object estimates with observed foreground and background probabilities $P_{f,b}$:
\begin{equation} \label{eq:dense_silhouette}
E_s = \int_{\Omega} H(\phi) P_{f}(x) + \left(1-H(\phi)\right) P_{b}(x) d\Omega,
\end{equation}
where $\phi$ is a 3D or 2D shape-kernel, and $H$ a mapping to a 2D foreground probability field, resembling an object mask of the current state.
Empirically, we found that 3D shape-kernels~\cite{2012-prisacariu-pwp3d} provide higher quality reconstructions when compared with a 2D formulation~\cite{2012-prisacariu} because more regions contribute to gradients.
While $H$ is a Heaviside function in the presence of 2D level-sets, we interpret signed distance samples of the DeepSDF volume as logits and compute a per-pixel foreground probability by accumulating samples along rays, similar to Prisacariu~\etal~\cite{2012-prisacariu}:
\begin{equation} \label{eq:dense_H}
H = 1 - \exp \prod_{\bx \text { on ray }} \left(1 - \text{sig}(\zeta \cdot
  \phi(\bx)) \right) \,,
\end{equation}
where $\zeta$ is a smoothing coefficient, 
and $1 - \text{sig}(\zeta \cdot \phi(\bx))$ the background probability at a sampling location $\bx$. A step-size of $\frac{r}{50}$ is chosen, where $r$ is the depth range of the object-space unit-cube.

\section{Experimental Evaluation}\label{sec:experiments}
Our focus is to evaluate the performance of FroDO on real-world datasets wherever possible. We evaluate \emph{quantitatively} in two scenarios:  \emph{(i)} single-view, single object on Pix3D~\cite{pix3d}; and \emph{(ii)} multi-view, single object on the \redwood{}~\cite{2016choiredwood} dataset. In  addition, we evaluate our full approach \emph{qualitatively} on challenging sequences from the real-world ScanNet dataset~\cite{dai2017scannet} that contain multiple object instances. 
In all cases we use MaskRCNN~\cite{he2017mask} to predict object detections and masks. We run Orb-SLAM~\cite{mur2015orb} to estimate trajectories and keypoints on Redwood-OS but use the provided camera poses and no keypoints on ScanNet.

\subsection{Single-View Object Reconstruction}
\begin{figure}[tb]
    \centering
    \includegraphics[width=0.95\linewidth]{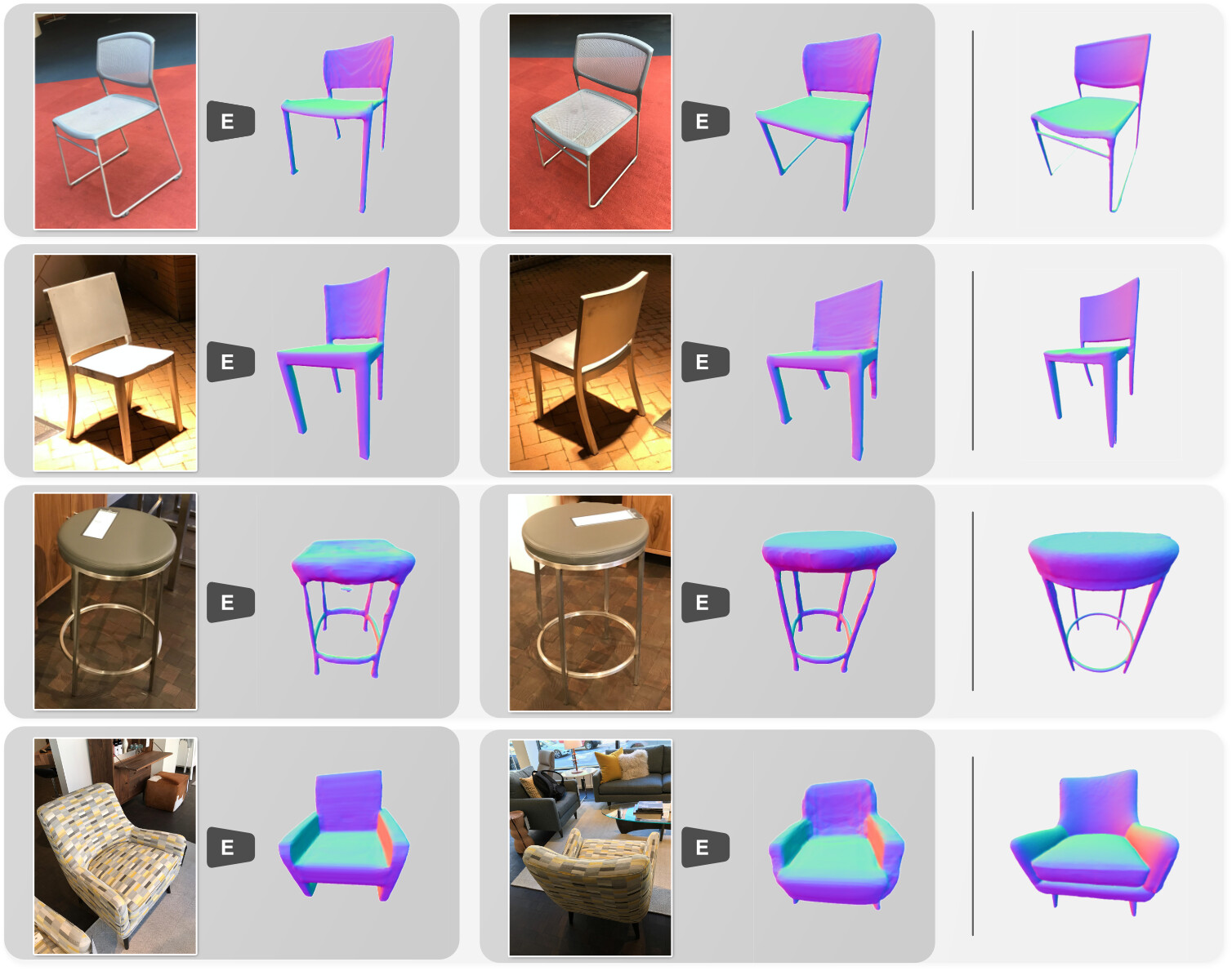}
    \caption{Examples of single view reconstruction on Pix3D dataset \cite{pix3d}. Ground truth on the right for reference.}
    \label{fig:pix3d}
\end{figure}
\begin{table}[tb]
  \begin{center}
  \small
    \begin{tabular}{l|c|c|c} % <-- Alignments: 1st column left, 2nd middle and 3rd right, with vertical lines in between
    \hline
       & IoU $\uparrow$ & EMD $\downarrow$ & CD $\downarrow$ \\
      \hline\hline
      3D-R2N2 \cite{3D-R2D2} & 0.136 & 0.211 & 0.239\\
      PSGN \cite{PSGN} & N/A & 0.216 & 0.200\\
      3D-VAE-GAN \cite{3D-VAE-GAN} & 0.171 & 0.176 & 0.182\\
      DRC \cite{DRC} & 0.265 & 0.144 & 0.160\\
      MarrNet \cite{2017-wu} & 0.231 & 0.136 & 0.144\\
      AtlasNet \cite{AtlasNet} & N/A & 0.128 & 0.125\\
      Sun et al. \cite{pix3d} & 0.282 & 0.118 & 0.119\\\hline
      Ours (DeepSDF Embedding) & \textbf{0.302} & \textbf{0.112} & \textbf{0.103}\\
      Ours (Joint Embedding)& \textbf{0.325} & \textbf{0.104} & \textbf{0.099}\\
      \hline
    \end{tabular}
    \caption{Results on Pix3D~\cite{pix3d}. Our method gives the highest Intersection over Union and lowest Earth Mover's and Chamfer Distances.}
    \label{tab:pix3d}
  \end{center}
  \vspace{-0.5cm}
\end{table}
First we evaluate the performance of our single-view shape code prediction network (Sec.~\ref{sec:method_singleviewshapeinference}) on the real world dataset Pix3D~\cite{pix3d}. Table~\ref{tab:pix3d} shows a comparison with competing approaches on the \emph{chair} category. The evaluation protocol described in \cite{pix3d} was used to compare IoU, Earth Mover Distance (EMD) and Chamfer Distance (CD) errors (results of competing methods are from \cite{pix3d}). Our proposed encoder network outperforms related work in all metrics. Table~\ref{tab:pix3d} also shows an improvement in performance when our new joint shape embedding is used (Ours Joint Embedding) instead of  DeepSDF~\cite{2019-park} (Ours DeepSDF Embedding). Figure~\ref{fig:pix3d} shows  example reconstructions.

\subsection{Multi-View Single Object Reconstruction}

We quantitatively evaluate our complete  multi-view pipeline on the \emph{chair} category of the real-world \redwood{} dataset~\cite{2016choiredwood} which contains single object scenes. We perform two experiments: an ablation study to motivate the choice of terms in the energy function (Table ~\ref{tab:ab_loss}) and a comparison of the performance of the different steps of our pipeline with related methods (Table~\ref{tab:ab_upgrade}). Table ~\ref{tab:shape360} includes a comparison of our dense photometric optimization with the two closest related approaches~\cite{2019-lin,liu2019dist} on a commonly-used synthetic dataset~\cite{2019-lin}.

\begin{table}
    \centering
    \small
    \begin{tabular}{c |l | c}
        % \toprule
        Optim. Method & Energy Terms & CD (cm.)\\
        \midrule
        Sparse & $E_s+ E_r$ \cmt{regularization + silhouette} & $8.97$ \\
        Sparse & $E_s+E_p+E_g+E_r$ \cmt{regularization + silhouette} & $8.59$ \\
        Sparse + Dense & $E_s+E_r$ \cmt{regularization + silhouette} & $7.41$ \\
        Sparse + Dense & $E_s+E_p+E_g+E_r$ \cmt{regularization+ silhouette + photo + slam keypoints} & $7.38$ \\
        \midrule
        % \bottomrule
    \end{tabular}
    \caption{Ablation study of estimates after sparse and dense optimization stages on the \redwood dataset. We compare the effect of different energy terms in Eq.~(\ref{eq:total_energy}).} 
    \label{tab:ab_loss}
    \vspace{-0.1cm}
\end{table}

\noindent \textbf{Ablation study.} Table~\ref{tab:ab_loss} shows an ablation study on different energy terms in our sparse and dense optimizations (Eq.~\ref{eq:total_energy}). The combination of geometric and photometric cues with a regularizer on the latent space achieves best results. The supplementary material includes further experiments on the effect of filtering object detections (Sec.~\ref{sec:method_detectionandassociation}) and the efficiency gains of using our joint embedding.

\begin{table}[t]
    \small
    \centering
    \begin{tabular}{l | c |c c c c}
        &  PMO {(o)} & PMO (r) & DIST (r) & Ours (r) \\
       \hline
       Cars & {0.661} & 1.187 & {0.919} & 1.202 \\
       Planes & {1.129} & 6.124 & 1.595 & {1.382} \\

    \end{tabular}
    \caption{Non-symmetric Chamfer distance (completion) on first 50 instances of the synthetic PMO~\cite{2019-lin} test set. While (o) indicates the original PMO method with its own initialization, (r) indicates random initialization.}
    \label{tab:shape360}
    \vspace{-0.2cm}
\end{table}

\noindent \textbf{Synthetic dataset.} Table~\ref{tab:shape360} shows a direct comparison of the performance on the synthetic PMO test set~\cite{2019-lin} of our dense optimization when only the photometric loss $E_p$ is used in our energy, with the two closest related methods: PMO~\cite{2019-lin} and DIST~\cite{liu2019dist}. Notably, both DIST and our approach achieve comparable results to PMO from only random initializations. When PMO is also initialized randomly the results degrade substantially. \todo{THIS NEEDS TO BE UPDATED Note that \methodtitle differs to DIST in that it decouples the number of required DeepSDF samples from the number of viewpoints used. At a resolution of 512x512, optimizing over 60 views DIST is therefore approximately 12 times slower then \methodtitle's dense optimization.}

\begin{figure*}[th]
    \centering
    \newcolumntype{Y}{>{\centering\arraybackslash}X}
    \begin{tabularx}{\linewidth}{@{}Y@{\,}Y@{\,}Y@{\,}Y@{\,}Y@{\,}Y@{}}
        % Row 1
        \includegraphics[width=\linewidth]{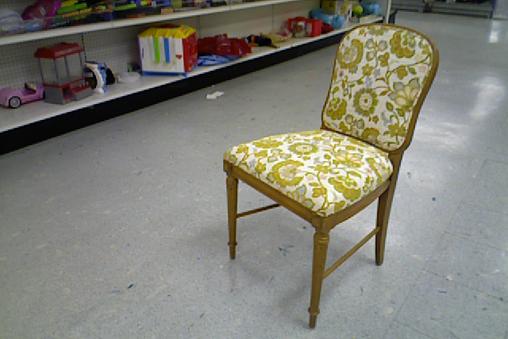}&
        \includegraphics[width=\linewidth]{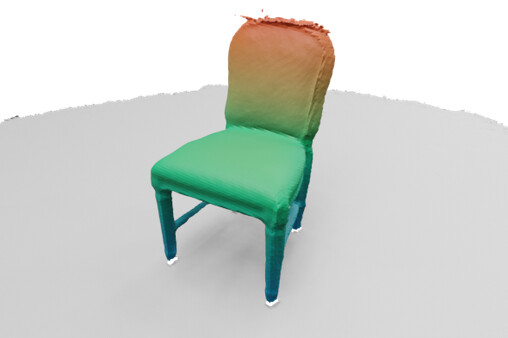}&
        \includegraphics[width=\linewidth]{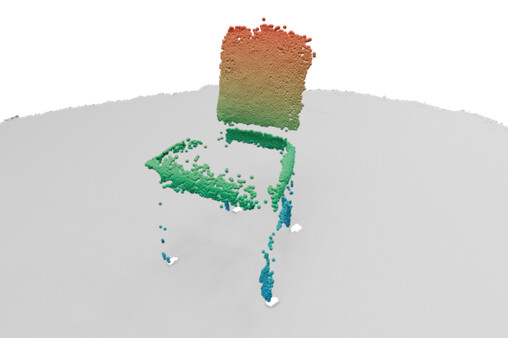}&
        \includegraphics[width=\linewidth]{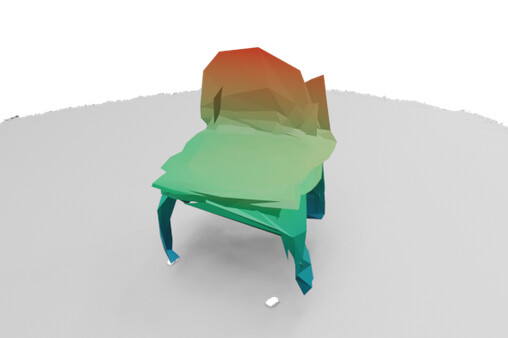}&
        \includegraphics[width=\linewidth]{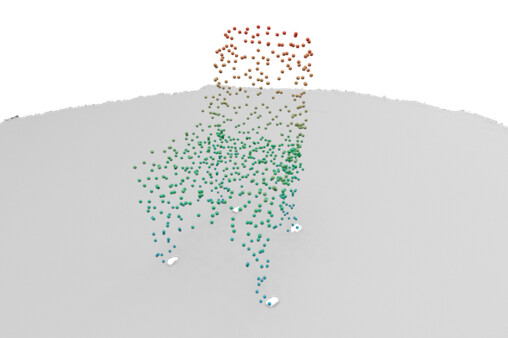}&
        \includegraphics[width=\linewidth]{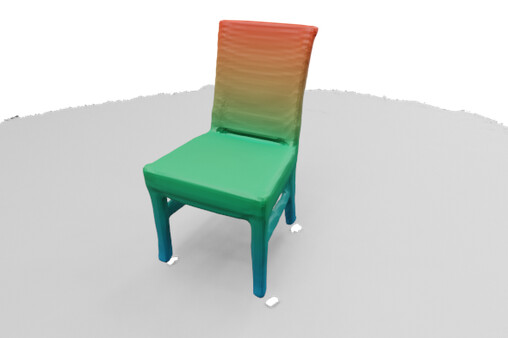}\\
        % Row 2
        \includegraphics[width=\linewidth]{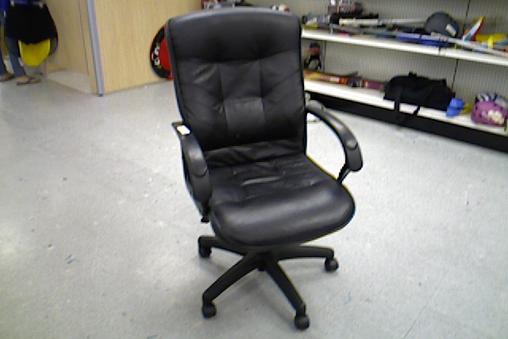}&
        \includegraphics[width=\linewidth]{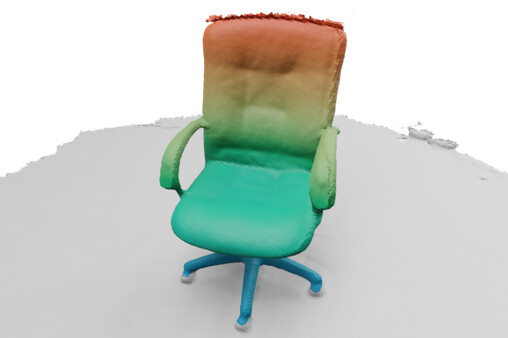}&
        \includegraphics[width=\linewidth]{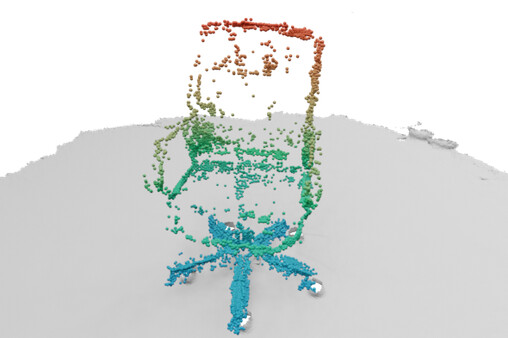}&
        \includegraphics[width=\linewidth]{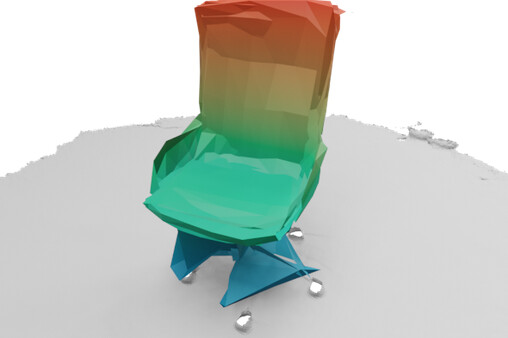}&
        \includegraphics[width=\linewidth]{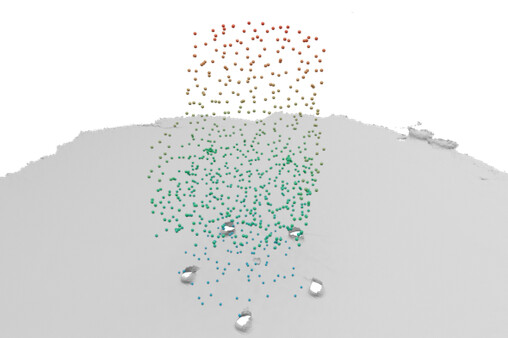}&
        \includegraphics[width=\linewidth]{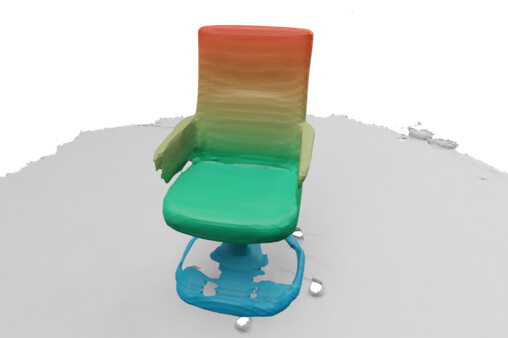}\\
        % Row 3
        \includegraphics[width=\linewidth]{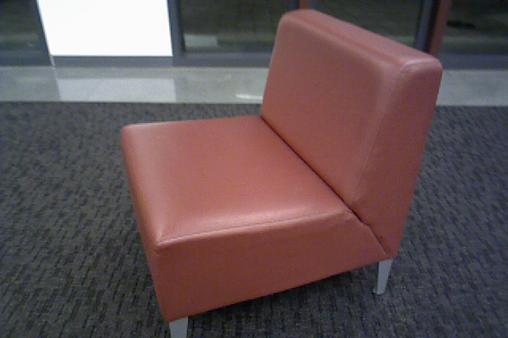}&
        \includegraphics[width=\linewidth]{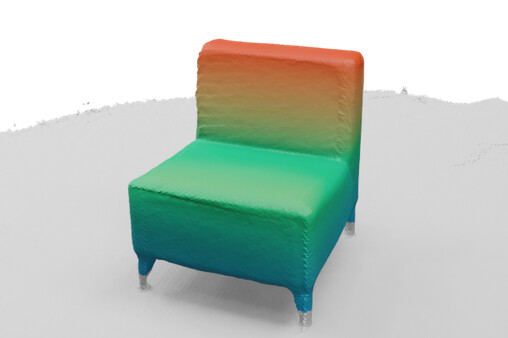}&
        \includegraphics[width=\linewidth]{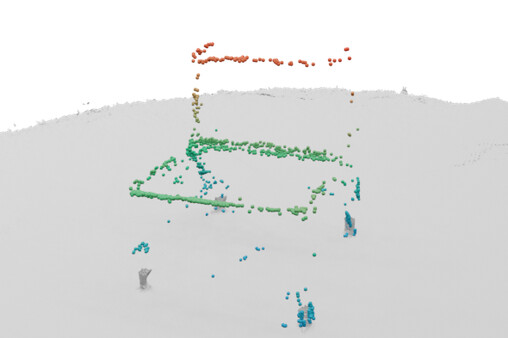}&
        \includegraphics[width=\linewidth]{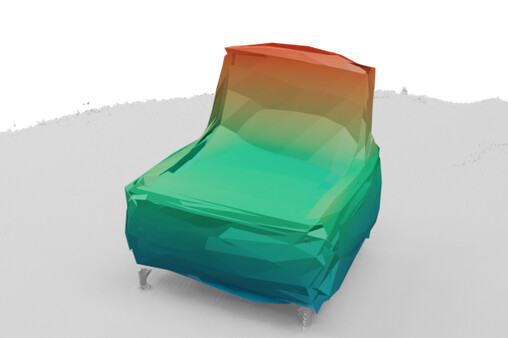}&
        \includegraphics[width=\linewidth]{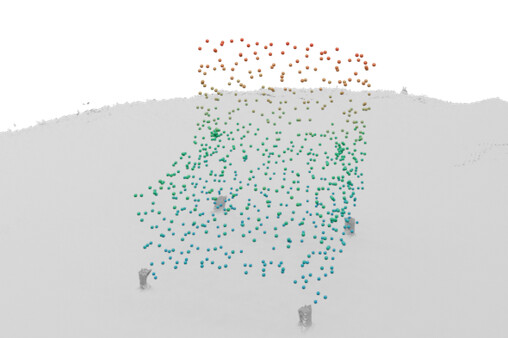}&
        \includegraphics[width=\linewidth]{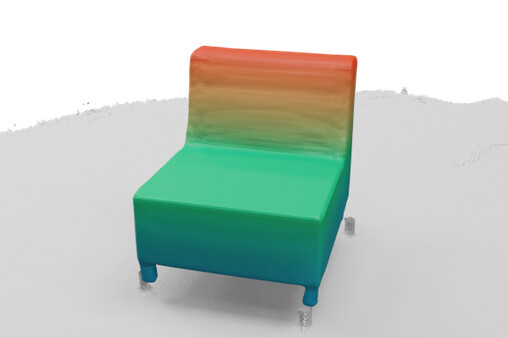}\\
         % Captions
         RGB &
         GT Scan &
         COLMAP &
         PMO &
         FroDO (sparse) &
         FroDO (dense)
    \end{tabularx}
    
    \caption{Example 3D reconstructions achieved with different approaches on three sample sequences from {\redwood}. In all cases 35 input views were used. 
\label{fig:redwood_result}}

\end{figure*}

\noindent \textbf{Redwood-OS dataset.}
Table~\ref{tab:ab_upgrade} shows a comparison with Pix2Vox~\cite{xie2019pix2vox}, a purely deep learning approach, and with PMO~\cite{2019-lin}, both of which are state-of-the-art. For reference, we also compare with COLMAP~\cite{schoenberger2016sfm,schoenberger2016mvs} a traditional SFM  approach. Since COLMAP reconstructs the full scene without segmenting objects, we only select points within the ground-truth 3D bounding box for evaluation. We report errors using: Chamfer distance (CD),  accuracy (ACC (5cm)), completion (COMP (5cm)) and F1 score  --  all four commonly used when evaluating on {\redwood}. Chamfer distance (CD) measures the symmetric error, while shape accuracy captures the 3D error as the distance between predicted points to their closest point in the ground truth shape and vice-versa in the case of shape completion. Both shape accuracy and completion are measured in percentage of points with an error below $5$cm. Following ~\cite{2019-lin}, we use an average of $35$ input frames sampled from the RGB sequences, though for completeness we show results with  $350$ views.   Fig.~\ref{fig:redwood_result} shows example reconstructions. 

We outperform Xie \etal~\cite{xie2019pix2vox} by a significant margin which could point to the lack of generalization of purely learning based approaches. We also outperform PMO~\cite{2019-lin}, a shape prior based optimization approach like ours, but which lacks our proposed coarse-to-fine shape upgrade. 
COLMAP fails to reconstruct full 3D shapes when the number of input images or the baseline of viewpoints is limited as it cannot leverage pre-learnt object priors. Although, as expected, the performance of COLMAP increases drastically with the number of input images, it requires hundreds of views to perform comparably to our approach.

\subsection{Multi-Object Reconstruction}
\label{sec:experiment_scannet}
\begin{figure*}[th]
    \centering
    \newcolumntype{Y}{>{\centering\arraybackslash}X}
    \begin{tabularx}{\linewidth}{@{}Y@{\,}Y@{\,}Y@{\,}Y@{}}
        % Row 1
        \includegraphics[width=\linewidth]{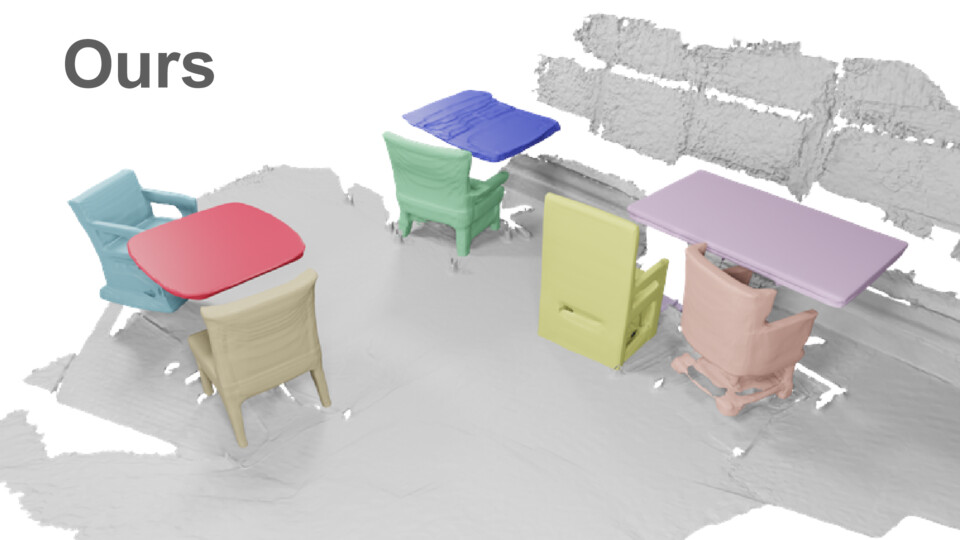}&
        \includegraphics[width=\linewidth]{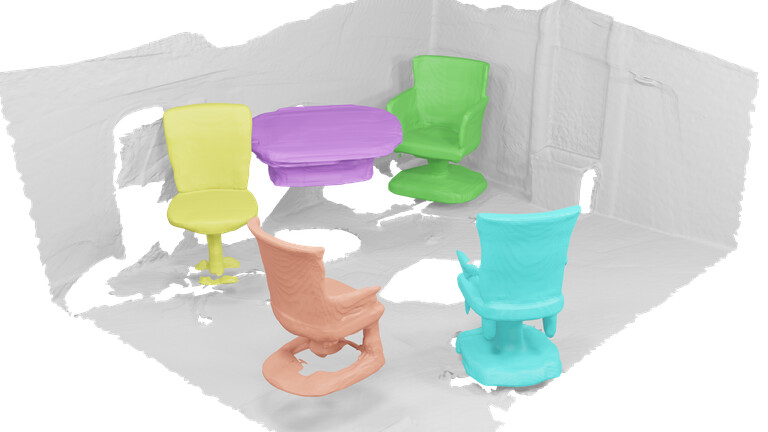}&
        \includegraphics[width=\linewidth]{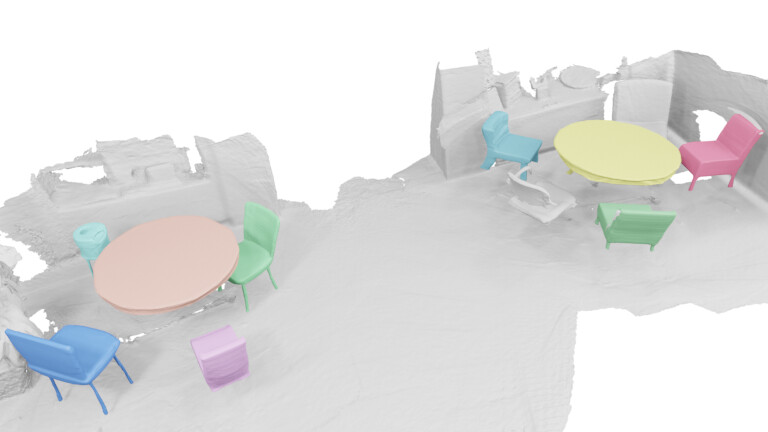}&
        \includegraphics[width=\linewidth]{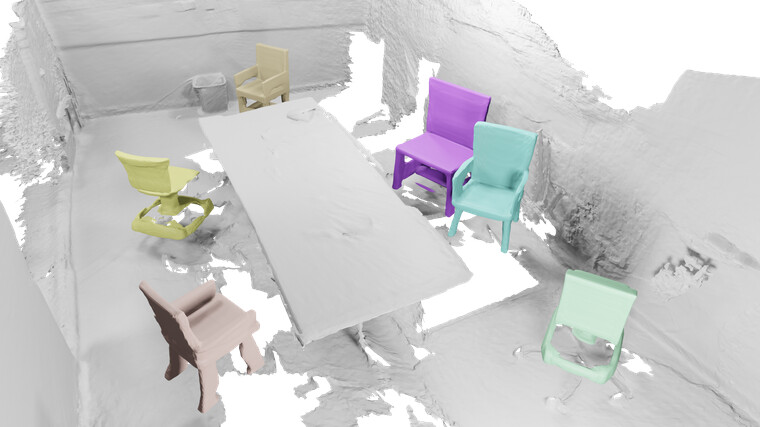}\\
        % Row 2, category:
        % \includegraphics[width=\linewidth]{figures/scannet/scene0609_00_gt.jpg}&
        % \includegraphics[width=\linewidth]{figures/scannet/scene0427_00_gt.jpg}&
        % \includegraphics[width=\linewidth]{figures/scannet/scene0355_00_gt.jpg}&
        % \includegraphics[width=\linewidth]{figures/scannet/scene0575_00_gt.jpg}&
        % Row 2, grayscale:
        \includegraphics[width=\linewidth]{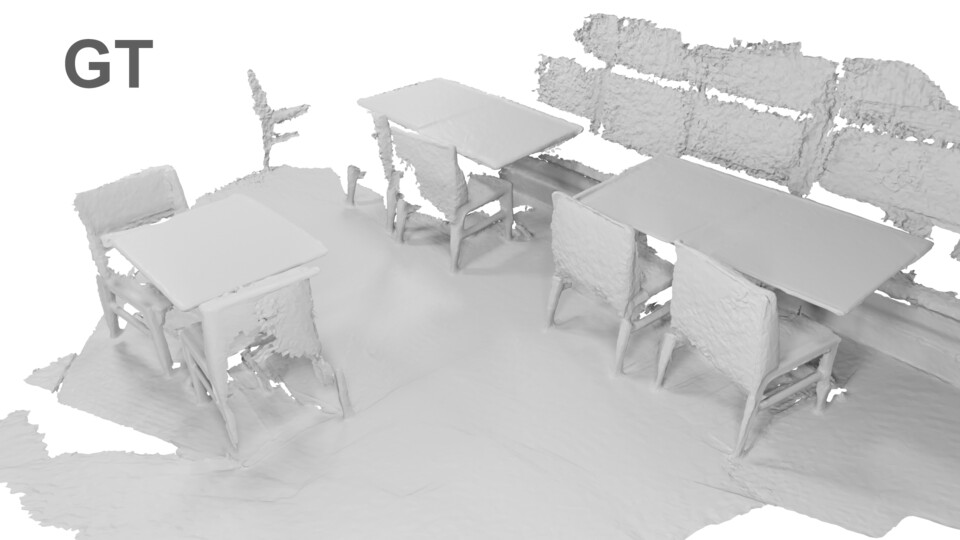}&
        \includegraphics[width=\linewidth]{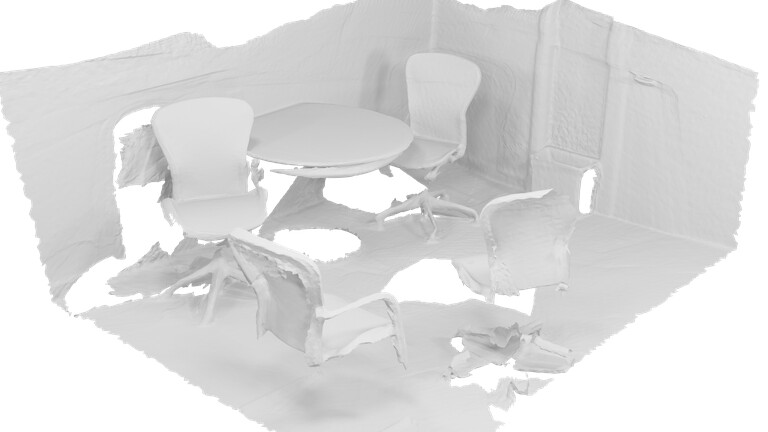}&
        \includegraphics[width=\linewidth]{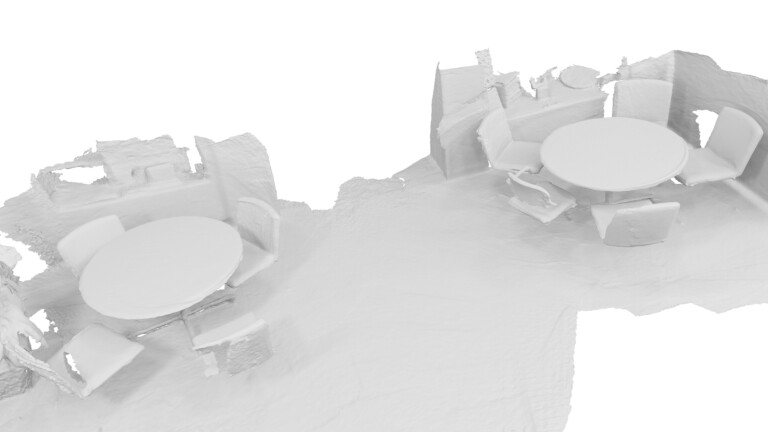}&
        \includegraphics[width=\linewidth]{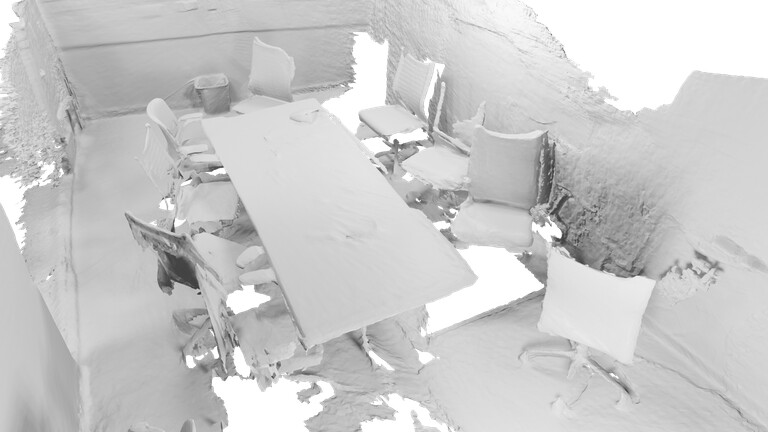}
    \end{tabularx}
    \caption{Qualitative results on four ScanNet RGB input sequences. We  reconstruct multiple instances of the chair and table classes. While outputs are satisfactory for the first three scenes, the last one highlights failures due to heavy occlusions and partial observations. Top row: Object instances reconstructed by FroDO are shown in colour while grey shows the ground truth background (not reconstructed by our method) for reference. Bottom row: full ground truth scan for comparison.}
	\label{fig:scannet_results}
\end{figure*}

We demonstrate qualitative results of our full approach on the ScanNet dataset~\cite{dai2017scannet} on challenging real world scenes with multiple object instances in Fig.~\ref{fig:teaser} and Fig.~\ref{fig:scannet_results}. MaskRCNN~\cite{he2017mask} was used to predict 2D b-boxes and masks. The association of object detections to 3D object instances becomes an additional challenge when dealing with multi-object scenarios. Our results show that our ray clustering approach successfully associates  detected bounding boxes across frames and our coarse-to-fine optimization scheme provides high quality object poses and reconstructions. 

\section{Conclusions and Discussion}\label{sec:discussion}
We introduced FroDO, a novel object-oriented 3D reconstruction framework that takes localized  monocular RGB images as input and infers the location, pose and accurate shape of the objects in the scene. Key to FroDO is the use of a new deep learnt shape encoding throughout the different shape estimation steps. We  demonstrated FroDO on challenging sequences from real-world datasets in single-view, multi-view and multi-object settings. An exciting open challenge would be to extend FroDO to the case of dynamic scenes with  independently moving objects.

\noindent \textbf{Acknowledgement}
The Univ. of Adelaide authors’ work has been supported by the Australian Research Council through the Centre of Excellence for Robotic Vision CE140100016 and Laureate Fellowship FL130100102, and UCL authors' work has been supported by the SecondHands project, funded from the EU Horizon 2020 Research and Innovation programme under GA No 643950.
\clearpage

\begin{table*}[t]
    \small
    \centering
    \setlength\tabcolsep{4pt}
    \begin{tabular}{l | cccccccc }
        % \toprule
        Method & \multicolumn{4}{c |}{Few observations (average 35 views)} & \multicolumn{4}{c}{Over-complete observations (average 350 views)} \\ 
        \midrule
         & CD (cm) & \makecell{ACC ($5$cm)} & \makecell{COMP ($5$cm)} & F1 score & CD (cm) & \makecell{ACC ($5$cm)} & \makecell{COMP ($5$cm)} & F1 score \\
        \midrule
        COLMAP~\cite{schoenberger2016sfm,schoenberger2016mvs} & 10.58 & \textbf{84.16} & 54.28 & 65.99 & \textbf{6.05} & \textbf{91.41} & \textbf{94.59} & \textbf{92.97}\\
        \midrule
        Pix2Vox~\cite{xie2019pix2vox} & 12.12 & 55.27 & 64.74 & 59.63 & 11.87 & 55.88 & 66.09 & 60.56 \\
        PMO~\cite{2019-lin} & 12.13 & 53.08 & 69.42 & 60.16 & 11.93 & 54.80 & 69.54 & 61.30\\
        \cmt{Meng's majority} \textbf{FroDO} Code Fusion (Vote) & 12.19 & 60.74 & 60.55 & 60.64 & 11.97 & 61.37 & 58.20 & 59.74 \\
        \cmt{Meng's} \textbf{FroDO} Code Fusion (Aver.) & 10.74 & 61.31 & 72.11 & 66.27 & 10.57 & 61.06 & 72.14 & 66.14  \\
        \textbf{FroDO} Optim. Sparse & 8.69 & 70.58 & 79.10 & 74.60 & 8.59 & 71.69 & 81.63 & 76.34 \\
        \textbf{FroDO} Optim. Dense & \textbf{7.38} & 73.70 & \textbf{80.85} & \textbf{76.64} & 7.37 & 74.78 & 81.08 & 77.32 \\
        \midrule
        % \bottomruae
    \end{tabular}
    \caption{Quantitative evaluation on 86 sequences of \redwood{}. We compare state of the art competitors Pix2Vox~\cite{xie2019pix2vox} and PMO~\cite{2019-lin} with the results at different stages of our multi-view pipeline (code fusion $\xrightarrow{}$ sparse optimization $\xrightarrow{}$ dense optimization). Average code outperforms majority voting. FroDo outperforms all methods when 35 input images are used.}
    \label{tab:ab_upgrade}
\end{table*}

\clearpage
{\small
\bibliographystyle{ieee}
\bibliography{main}
}

\end{document}

% https://tex.stackexchange.com/questions/55764/input-a-figure-between-title-and-body-in-twocolumn-form
\twocolumn[{%
    \renewcommand\twocolumn[1][]{#1}%
    \maketitle
    \centering
    \vspace{-1.5cm}
    \includegraphics[width=0.24\linewidth]{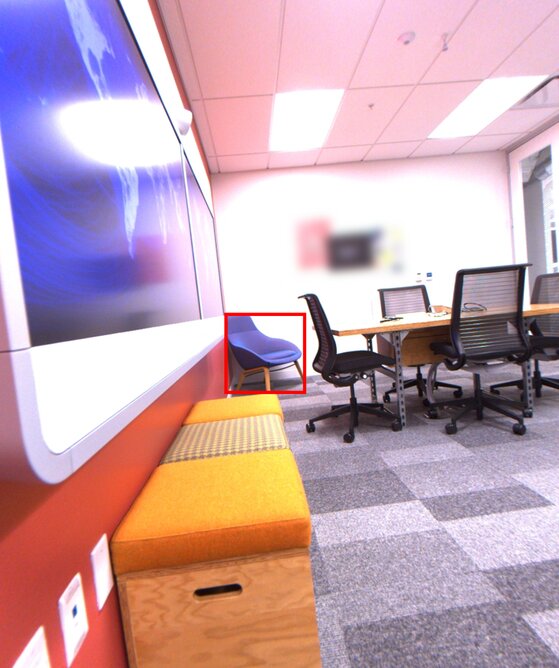}
\includegraphics[width=0.24\linewidth]{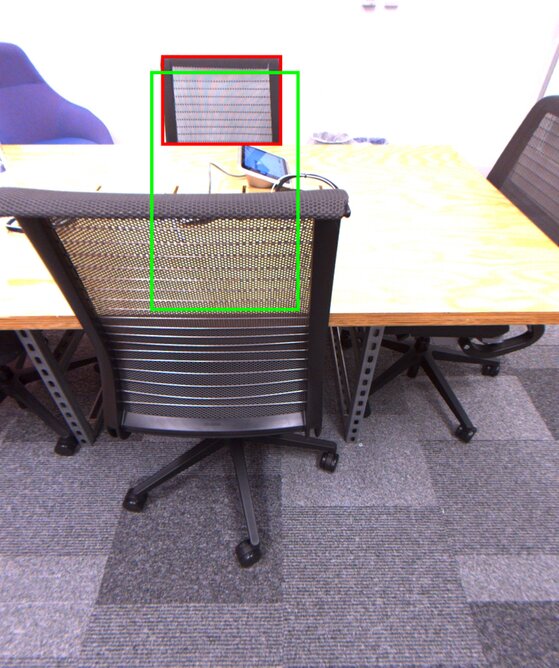}
\includegraphics[width=0.24\linewidth]{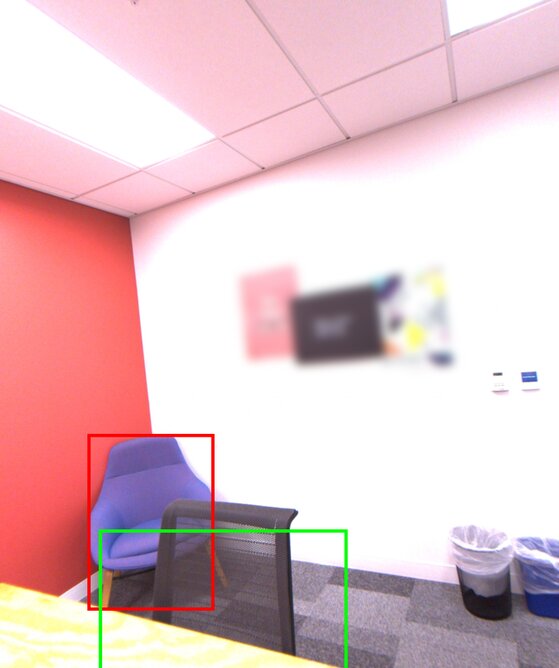}
\includegraphics[width=0.24\linewidth]{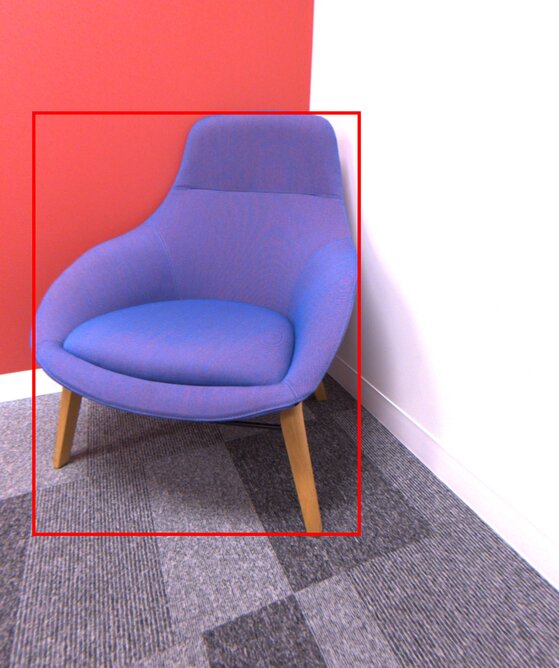}
\captionof{figure}{Examples of 2D detections filtering based on the projection of 3D bounding box. The red and green boxes are 2D detections and projections of 3D bounding boxes respectively. From left to right: reject due to size, reject due to occlusion, reject due to overlap, accept.\label{fig:supp_detection_prune}}
    \vspace{1cm}
}]

\section{Detection Pruning}
After estimating the bounding box of an object, good views are selected by employing a pruning scheme. This scheme uses three criteria to reject frames based on (1) bounding box size, (2) occlusions and (3) overlap with other objects. (1) is implemented via a threshold on the width and height of the bounding box (BB), and (2,3) are implemented using a threshold on the intersection over union (IoU) of the bounding boxes. 
Figure~\ref{fig:supp_detection_prune} illustrates these strategies. The selection helps to identify better initial detections to extract good shape codes and later on leads to an optimization with clean energy terms. An ablation study on the detection pruning on \redwood{} dataset is demonstrated in Table~\ref{tab:ab_filter}.

\section{DP-mean based Line Segment Clustering}
Dirichlet Process (DP)-mean clustering is similar to K-mean clustering as it also runs in an Expectation-Maximization manner.  A key difference is that the total number of cluster is unknown at first. A new cluster is generated when the distance between a line segment and any existing clusters is larger than a cluster penalty threshold, which we set to $0.4$. Our clustering algorithm runs as follows: Given a new object detection, which is represented by a line segment, we calculate the distance between this new line segment and existing clusters. If the distance is larger than the cluster penalty threshold, this observation becomes a new cluster, the cluster center is initialized at the median point of the line segment. Algorithm~\ref{algo:dp-mean} details each step.
\begin{algorithm}
\SetAlgoLined
\KwIn{$\boldsymbol{r_1}, \boldsymbol{r_2}, ..., \boldsymbol{r_n}: n \text{ rays} \linebreak \lambda: \text{ cluster penalty threshold}$}
\KwOut{$c_1, c_2, ..., c_m: m \text{ object clusters}$}
1. init. $\boldsymbol{\mu_1} = mid(\boldsymbol{r_1})$, $mid()$ is to compute the middle point of a ray\;
2. \While{not converged}{
    \For{each $\boldsymbol{r_i}$}{
    $d_{ij} = \min\limits_{\mu_1\dots k} dist(\boldsymbol{\mu_k}, \boldsymbol{r_i})$\;
    \eIf{$d_{ij} \ge \lambda$}{
        set k = k + 1\;
        $\boldsymbol{\mu_k}=mid(\boldsymbol{r_i})$\;
        }
        {$z_i=j$\;}
    }
    $k$ clusters are generated, where $c_k=\{\boldsymbol{r_i} | z_i=k\}$\;
    \For{each $c_i$}{
        $\boldsymbol{\mu_i} = lstsq(c_k)$\;
    }
 }
 \caption{DP-mean clustering for 3D line segments}
 \label{algo:dp-mean}
\end{algorithm}

\begin{table}[t]
    \centering
    \begin{tabular}{l | c | c }
        % \toprule
        Method & Vote & Average \\
        \midrule
        CD [cm] w/o occlusion filter & $12.17$ & $11.73$ \\
        CD [cm] w/ occlusion filter & $11.97$ & $10.57$ \\
        \midrule
        % \bottomruae
    \end{tabular}
    \caption{Ablation study on using the inferred 3D bounding box to filter occluded views.}
    \label{tab:ab_filter}
\end{table}

\section{Dense Shape Code Optimization}

In order to implement the dense optimization efficiently, certain measures were taken to achieve an effective optimization procedure. The following two sections explain decisions for the formulation of analytical partial derivatives and for the implementation these.

\subsection{Analytical Partial Derivatives}

As the DeepSDF decoder yields signed distance values, in contrast to the pointcloud decoder which outputs object coordinates, neither $\energyg$ nor $\energyp$ are explicitly defined. Therefore, dense object coordinates are extracted by sampling the zero-crossing of the distance field and Jacobians are derived analytically. While most of the derivatives of $\energyp$ and $\energyg$ wrt. code and pose follow the chain rule, the relevant term $\pdv{\bx}{\bz} = \pdv{\bx}{\Gd} \pdv{\Gd}{\bz}$ needs more attention.

\begin{figure}
    \begin{center}
        \includegraphics[width=.8\linewidth]{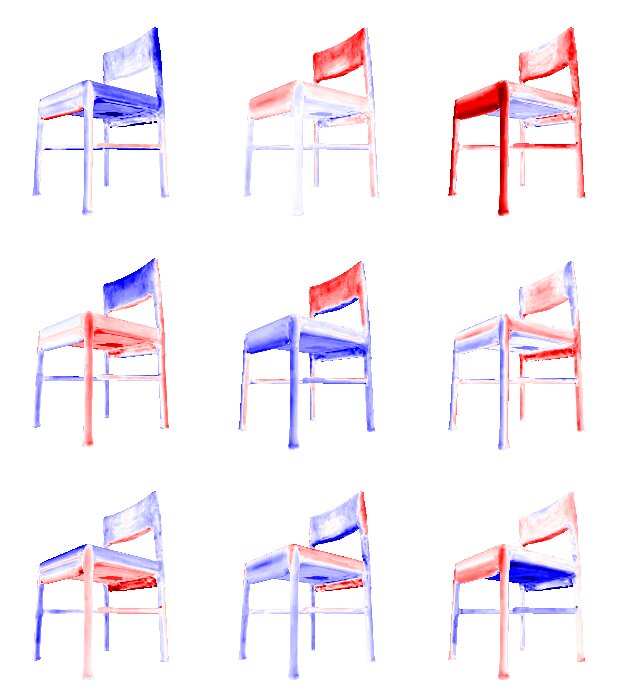}
    \end{center}
    \caption{Visualization of $\pdv{\Gd}{\mathbf{z}}$ for 9 different code components, demonstrating shape changes when single components change. Red (positive values) indicates an intrusion, while blue (negative values) indicates an extrusion.}
    \label{fig:dsdf_dcode}
\end{figure}

Intuitively, the change of a surface coordinate $\bx$ with respect to a change in code $\bz$ is ambiguous due to potential changes in topology. However, as shown by~\cite{atzmon2019controlling} it is possible to derive a first-order approximation as follows. Given a location $\bx(\bz)$ on the surface as a function of code $\bz$, the corresponding distance value remains constant at the zero-crossing, implying:
\begin{equation}
\pdv{\Gd(\bx(\bz), \bz)}{\bz} = 0
\end{equation}
Using the multivariable chain rule, and solving for $\pdv{\mathbf{x}}{\Gd}$, yields:
\begin{equation}
    \pdv{\bx}{\Gd} = -\pdv{\Gd}{\bx}^{-1} \pdv{\Gd}{\mathbf{z}} 
\end{equation}
Here, the pseudo-inverse of $\pdv{\Gd}{\bx}$ is orthogonal to the tangent plane and hence points in the direction of the surface normal.
Since the SDF decoder $\Gd$ is differentiable, $\pdv{\mathbf{x}}{\mathbf{z}}$ can be calculated easily using this first-order approximation. 
Figure~\ref{fig:dsdf_dcode} visualizes $\pdv{\Gd}{\mathbf{z}}$ for 9 different shape code components and demonstrates that many parts of an object are affected, when a single component changes.

\subsection{Implementation Details}
Multiple techniques are employed in order to speed up the dense optimization stage. Since executing the network is the most expensive step in the optimization, the number of forward and backward passes are kept to a minimum. At the beginning of each iteration the DeepSDF volume is sampled with an adaptive resolution. First, on a dense grid with a distance of $\delta$ between samples along a every dimension and afterwards at a higher resolution for voxels close the object boundary, i.e. when $-\delta < \Gd < \delta$, where $\delta$ is the DeepSDF truncation factor as described by Park~\etal~\cite{2019-park}. A mesh corresponding to the current code estimate is then obtained by running marching cubes~\cite{1995-chernyaev} on the samples and used as a proxy for the current state. Each vertex contains $\pdv{\Gd}{\mathbf{x}}$ and $\pdv{\Gd}{\mathbf{z}}$ and as a result data required to minimize the energies is generated by rendering the mesh to observed detections and is independent of sampling. Rasterized mesh data is directly copied from OpenGL to CUDA for the optimization.

Further, a pyramid scheme is implemented which aids convergence when starting from a bad initialization, improves run-time and reduces the memory footprint of the optimization. For every detection with width $w$ and height $h$, only pyramid levels $l$ with a bounded shape ($r_{min}^2 < \frac{w \cdot h}{4^{l}} < r_{max}^2$) are considered. In our experiments, $r_{min}$ and $r_{max}$ are always $40$ and $400$ respectively.

\section{Network Architecture and  Training}\label{sec:supp_network}
\subsection{Decoder for Joint embedding} 
\label{sec:decoder}
The shared joint embedding is $64$ dimensional. The network is split into two independent branches for point cloud and DeepSDF respectively. The point cloud branch is composed of four fully connected layers each with $512$, $1024$, $2048$, $2048\times3$ as output dimensions. We use ReLU as a nonlinear activation function following each fully connected layer except the last one. Readers can refer to \cite{2019-park} for a detailed discussion of the DeepSDF architecture that is employed in the dense branch.
The joint autodecoder is trained for $2000$ epochs with a batch size of $64$. There are $16384$ SDF samples for each shape. The learning rate for network parameters and latent vectors are $5 \cdot 10^{-3}$, and $10^{-3}$ respectively, decayed by $0.5$ for every $500$ epochs.

\subsection{Encoder Network}
\label{encoder}
After we train the decoder network as described in Sec. \ref{sec:decoder}, we obtain a set of embeddings $\mathbf{z}\in\mathcal{R}^{64}$ for the corresponding CAD models. In the second stage, we train an encoder network that maps an image to the latent code. We tailor ResNet50 to output a vector of dimension 64 and initialize the network with pretrained models. We train the network for 50 epochs to minimize a Huber loss with a polynomial decaying learning rate of $10^{-3}$. The network for the embedding is trained in a way similar to Li \textit{et al.}~\cite{li2015joint}. However, our deep learning based shape embedding is very different from the non-parametric embedding used in \cite{li2015joint}.

\begin{figure*}[thp]
    \begin{center}
        \includegraphics[width=.8\linewidth]{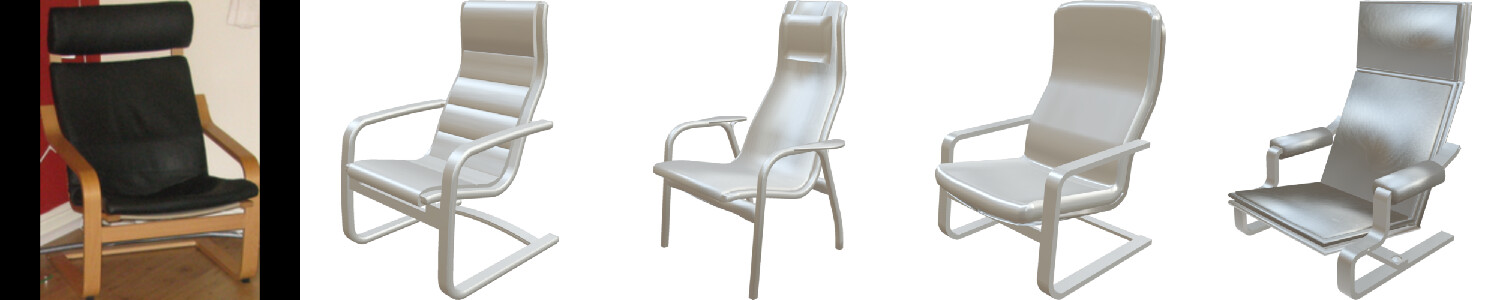}
        \\
        \includegraphics[width=.8\linewidth]{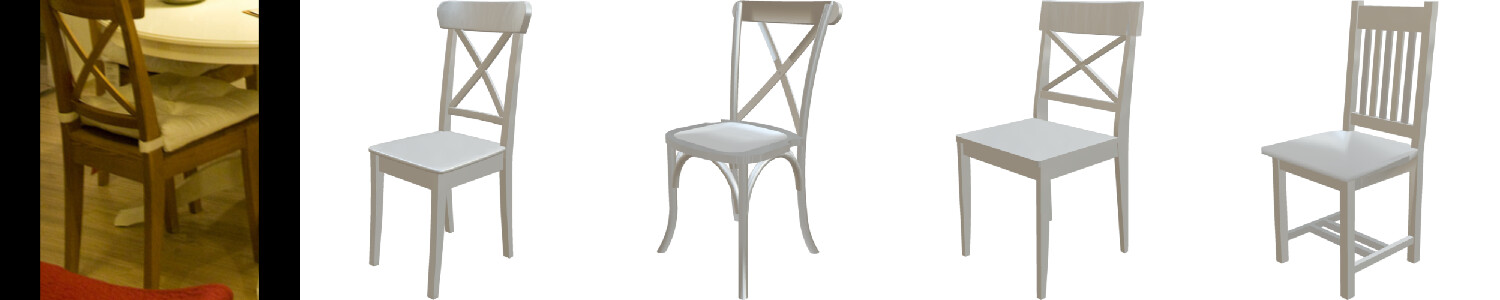}
        %\\
        %\includegraphics[width=.8\linewidth]{figures/pix3d/pix3d_retrieval_3.jpg}
        \\
        \includegraphics[width=.8\linewidth]{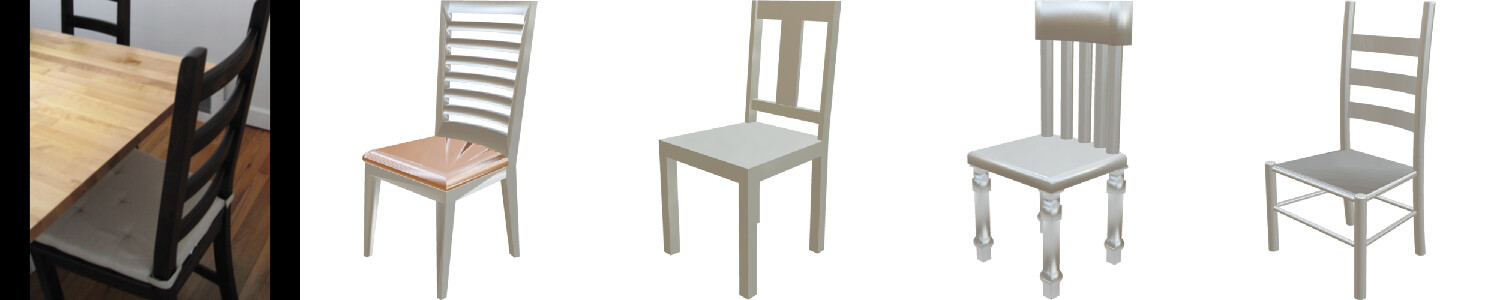}
        %\\
        %\includegraphics[width=.8\linewidth]{figures/pix3d/pix3d_retrieval_5.jpg}
        \\
        \includegraphics[width=.8\linewidth]{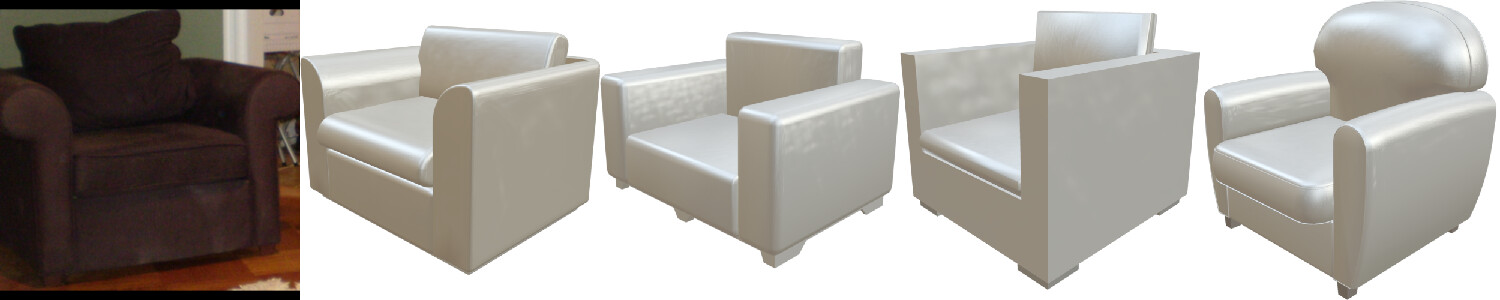}
        \\
        \includegraphics[width=.8\linewidth]{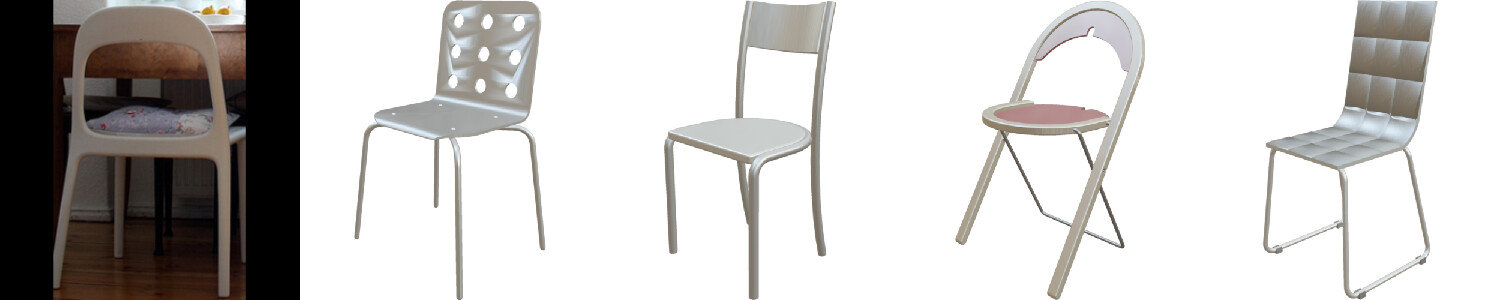}
        
    \end{center}
    \caption{From left to right: input image and the nearest CAD models in latent space sorted by distances.}
    \label{fig:pix3d_retrieval}
\end{figure*}

\section{CAD retrieval on Pix3D dataset}

In Fig. \ref{fig:pix3d_retrieval} we show more examples of CAD model retrieval on Pix3D dataset \cite{pix3d}. We find the nearest CAD model based on Euclidean distance of our shape codes.
% }

\section{Dual-Representation Efficiency Gain}\label{sec:supp_speed}
One motivation for using a shared code space for a point-based and SDF-based representation is that the efficiency of a sparse optimization can be leveraged, while exploiting richer information when subsequently applying fewer dense iterations. A comparison of optimization run-times shows that the pointcloud-based representation is approximately two orders of magnitude faster than the mesh optimization used by PMO~\cite{2019-lin} and DeepSDF~\cite{2019-park} as shown in Table~\ref{tab:ab_speed}.
\begin{table}[tbp]
    \centering
    \small
    \begin{tabular}{l|c|c}
        % \toprule
        Method & sec. /iteration & \# of iteration\\
        \midrule
        PMO~\cite{2019-lin} & 12.59 & 100 \\
        \hline
        Optim. Dense & 4.96 & 100 \\
        Optim. Sparse & 0.07 & 200 \\
        \midrule
        % \bottomrule
    \end{tabular}
    \caption{Speed comparison between PMO~\cite{2019-lin} and our method on the same sequence of 60 frames.}
    \label{tab:ab_speed}
\end{table}

% \section{Optimizing shared object codes}\label{sec:supp_shared_codes}

% In some scenarios it is common that identical objects exist multiple times, such as office chairs in professional environments. This can be exploited by optimizing a single code for multiple instances of the same object while optimizing individual poses, which is especially useful when only partial observations for each instance exist. Identical objects can be identified based on the distance in code space, as visualized in Figure~\ref{fig:instance_similarity} that shows pairwise similarities of objects.

\section{Redwood Dataset Augmentation}
While Pix3d~\cite{pix3d} is a real-world dataset available for single-view object shape reconstruction, multi-view datasets with ground truth 3D data are scarce. Previous multi-view learning based methods primarily evaluate their methods on synthetic data~\cite{2019-lin}. 
We post-process the \redwood{} dataset~\cite{2016choiredwood} and will release it to facilitate benchmarking on real-world data. It contains RGBD sequences and dense full scene reconstructions. To isolate object shapes for evaluation, we manually select sequences that have full 3D object shapes and manually segment them from the full scene mesh. We run RGBD ORB-SLAM2~\cite{murORB2} to estimate camera poses. 

\section{Qualitative results on \redwood{} dataset}\label{sec:supp_redwood_result}
\begin{figure*}[th]
    \centering
    \begin{subfigure}[t]{\textwidth}
        \centering
        \includegraphics[width=0.19\columnwidth]{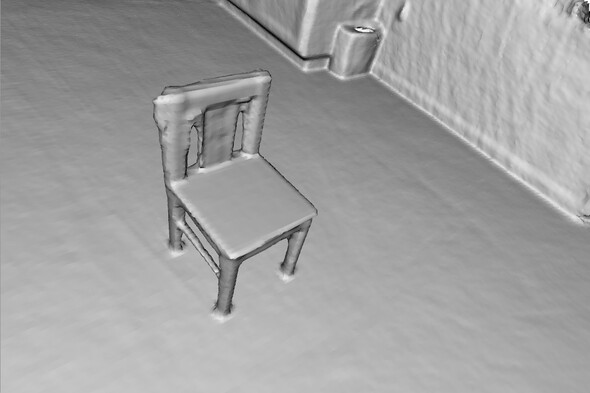}
        \hfill
        \includegraphics[width=0.19\columnwidth]{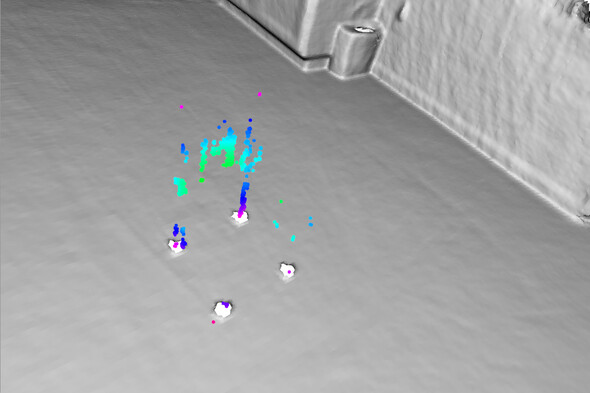}
        \hfill
        \includegraphics[width=0.19\columnwidth]{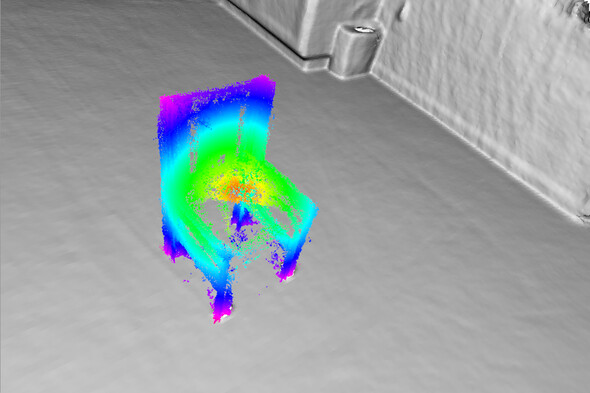}
        \hfill
        \includegraphics[width=0.19\columnwidth]{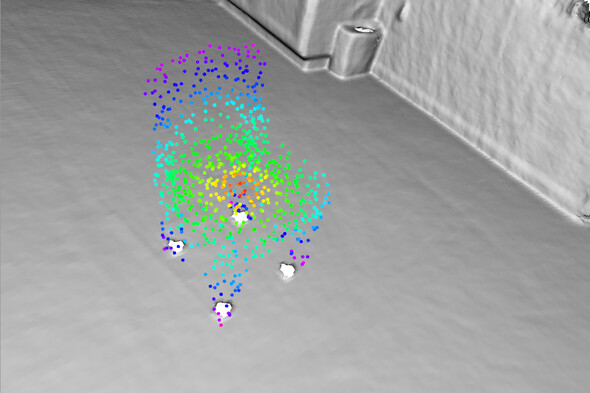}
        \hfill
        \includegraphics[width=0.19\columnwidth]{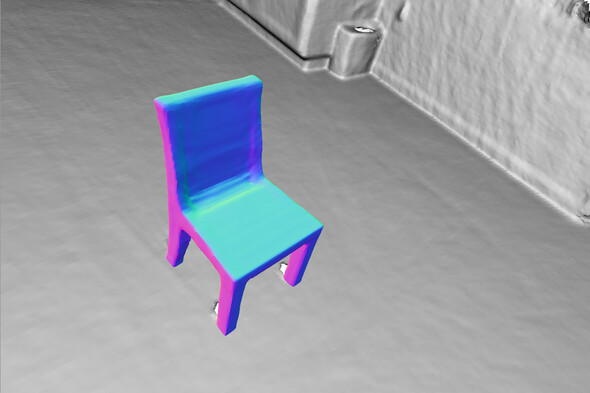}
    \end{subfigure}

    \begin{subfigure}[t]{\textwidth}
        \centering
        \includegraphics[width=0.19\columnwidth]{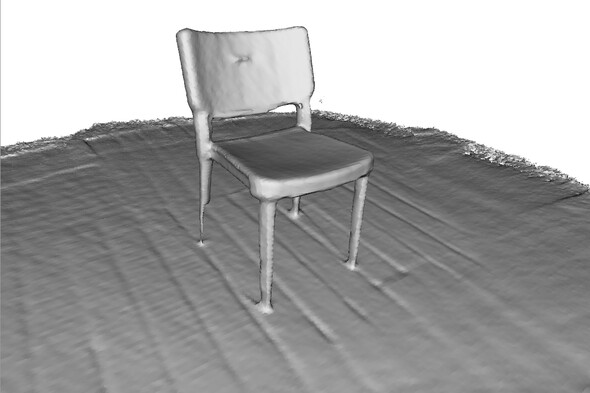}
        \hfill
        \includegraphics[width=0.19\columnwidth]{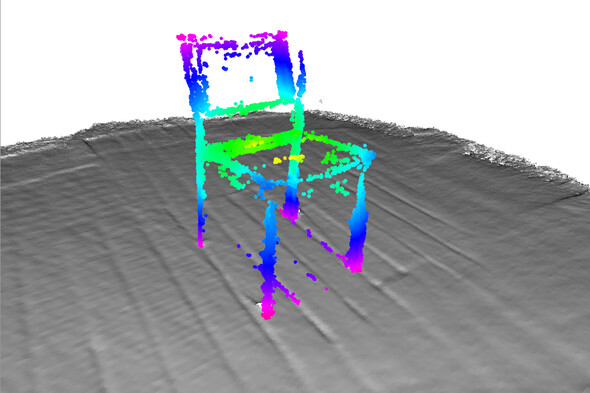}
        \hfill
        \includegraphics[width=0.19\columnwidth]{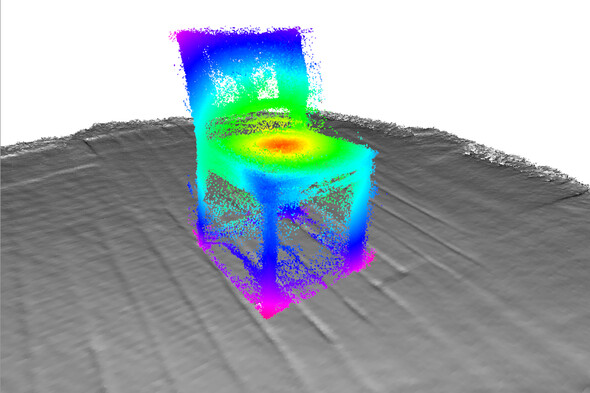}
        \hfill
        \includegraphics[width=0.19\columnwidth]{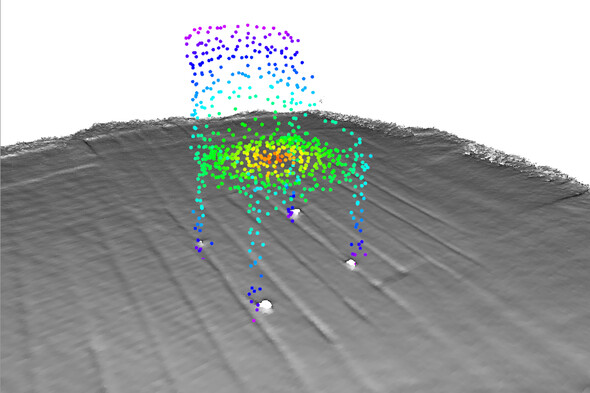}
        \hfill
        \includegraphics[width=0.19\columnwidth]{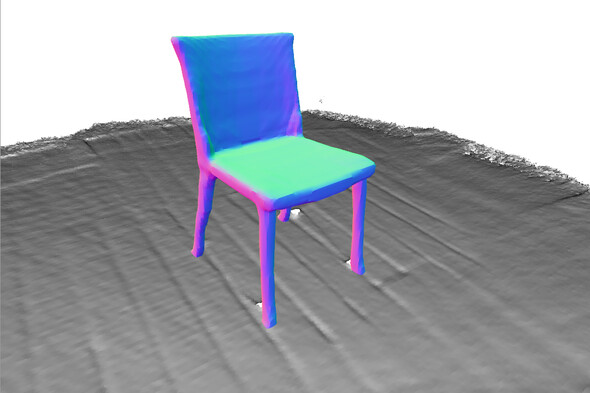}
    \end{subfigure}

    \begin{subfigure}[t]{\textwidth}
        \centering
        \includegraphics[width=0.19\columnwidth]{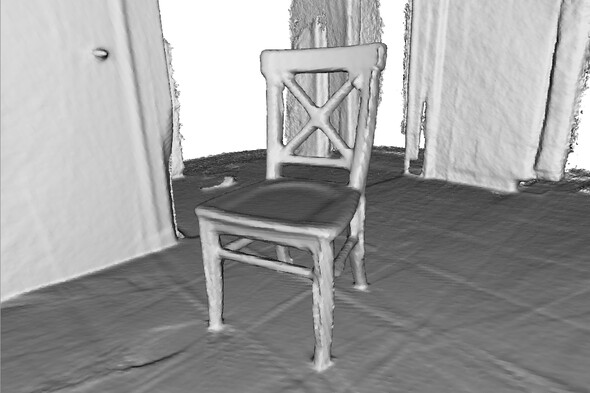}
        \hfill
        \includegraphics[width=0.19\columnwidth]{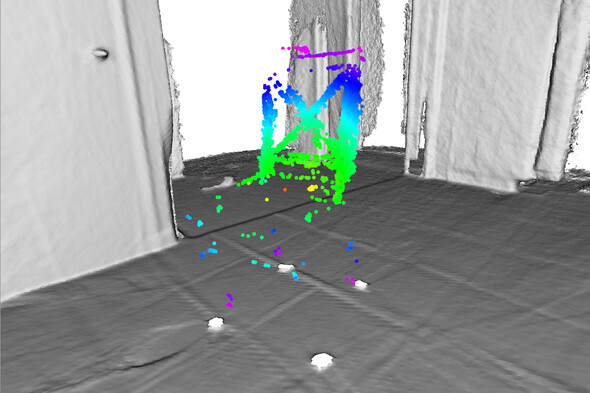}
        \hfill
        \includegraphics[width=0.19\columnwidth]{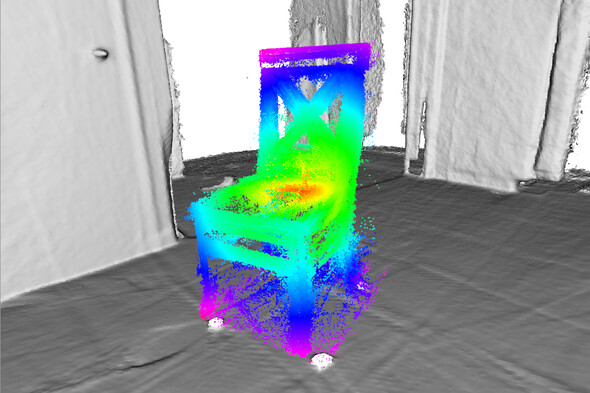}
        \hfill
        \includegraphics[width=0.19\columnwidth]{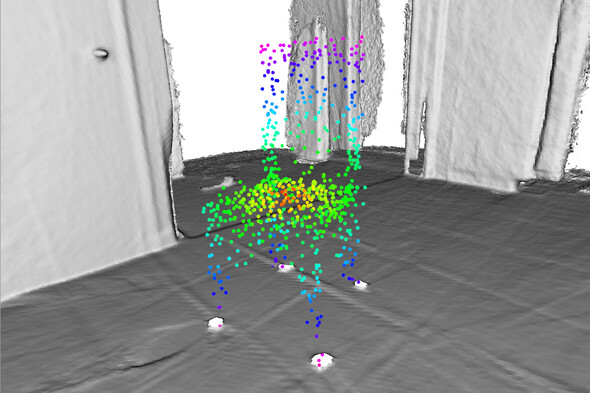}
        \hfill
        \includegraphics[width=0.19\columnwidth]{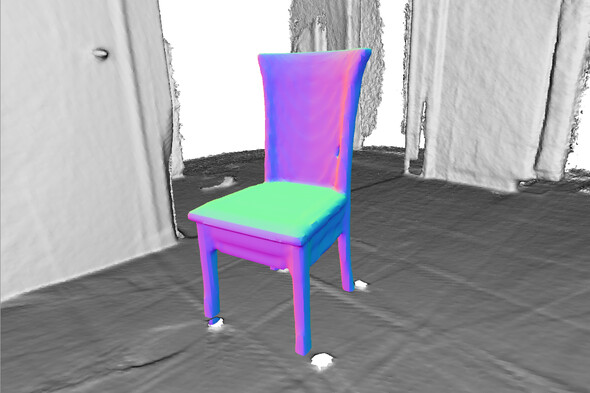}
    \end{subfigure}

    \begin{subfigure}[t]{\textwidth}
        \centering
        \includegraphics[width=0.19\columnwidth]{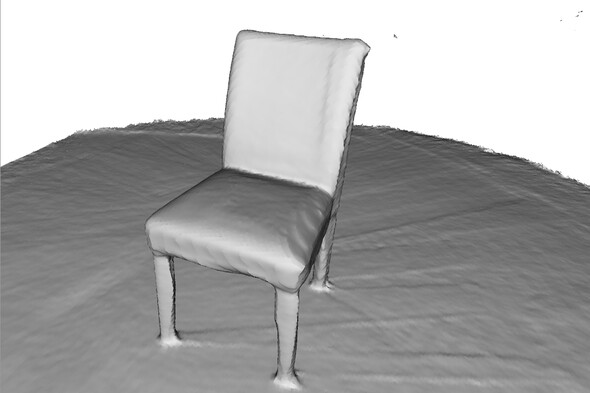}
        \hfill
        \includegraphics[width=0.19\columnwidth]{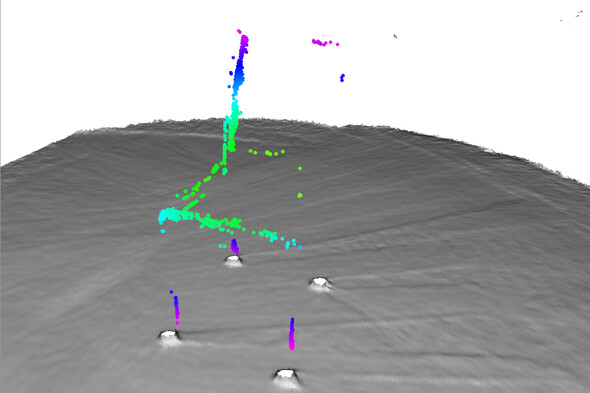}
        \hfill
        \includegraphics[width=0.19\columnwidth]{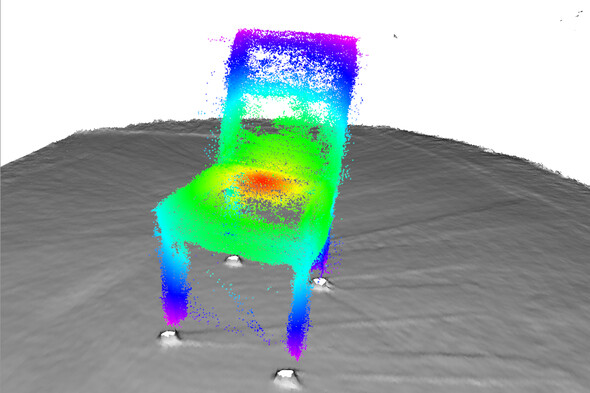}
        \hfill
        \includegraphics[width=0.19\columnwidth]{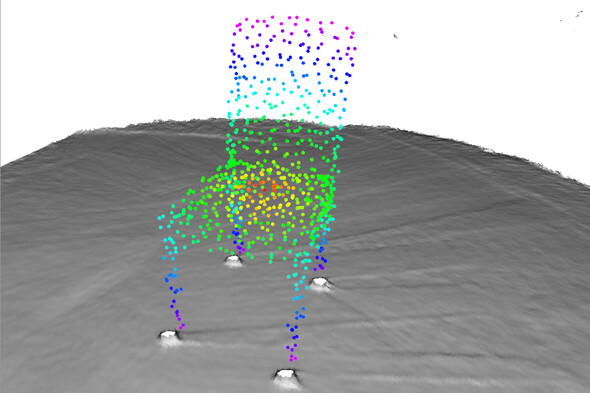}
        \hfill
        \includegraphics[width=0.19\columnwidth]{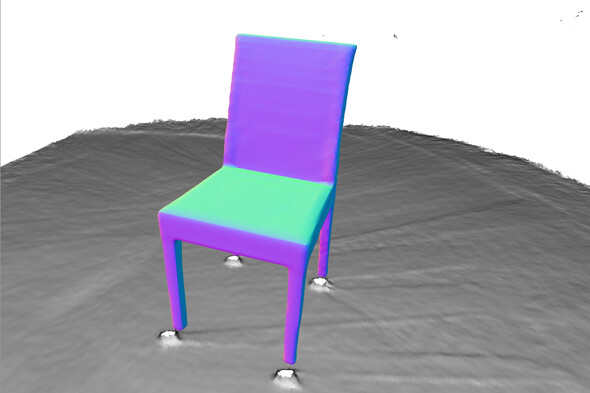}
    \end{subfigure}
    
    %%%% new images
    \begin{subfigure}[t]{\textwidth}
        \centering
        \includegraphics[width=0.19\columnwidth]{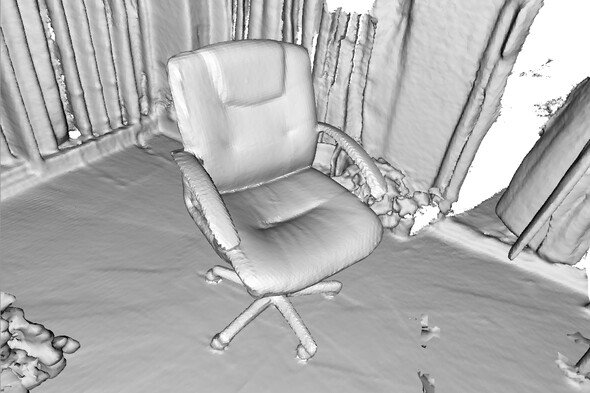}
        \hfill
        \includegraphics[width=0.19\columnwidth]{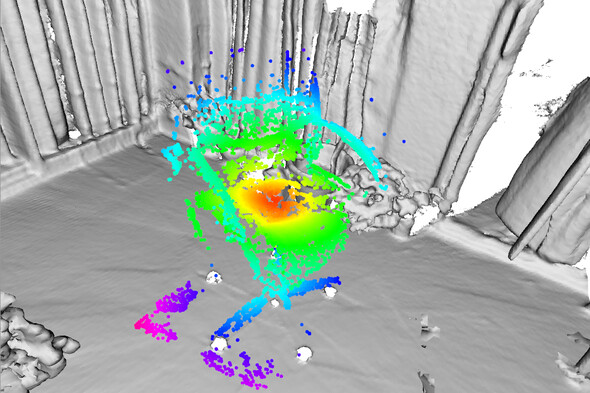}
        \hfill
        \includegraphics[width=0.19\columnwidth]{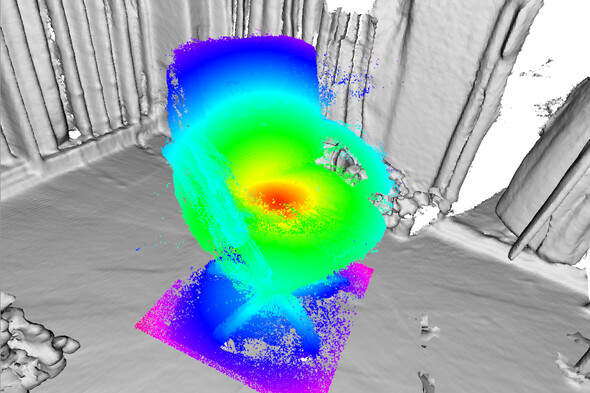}
        \hfill
        \includegraphics[width=0.19\columnwidth]{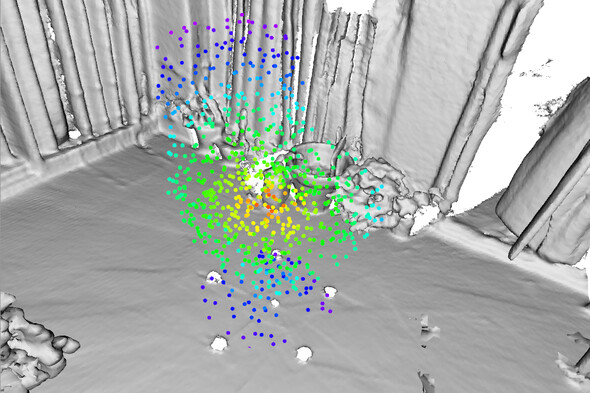}
        \hfill
        \includegraphics[width=0.19\columnwidth]{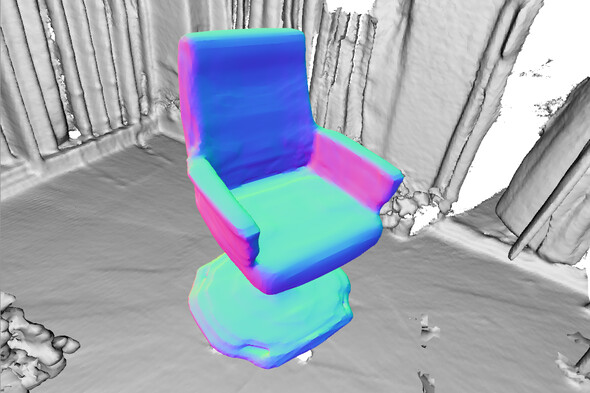}
    \end{subfigure}
    
    \begin{subfigure}[t]{\textwidth}
        \centering
        \includegraphics[width=0.19\columnwidth]{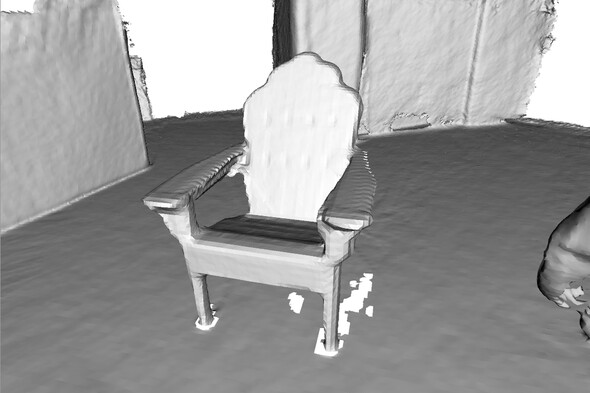}
        \hfill
        \includegraphics[width=0.19\columnwidth]{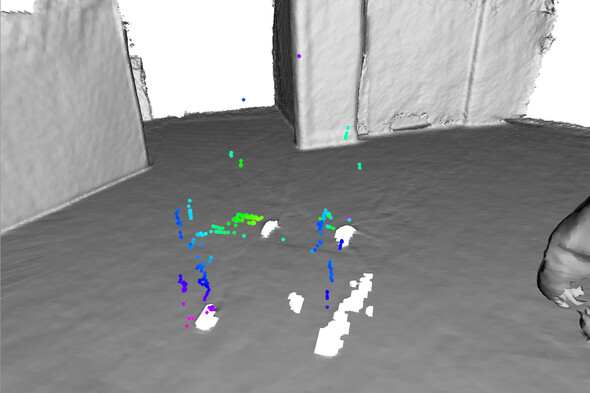}
        \hfill
        \includegraphics[width=0.19\columnwidth]{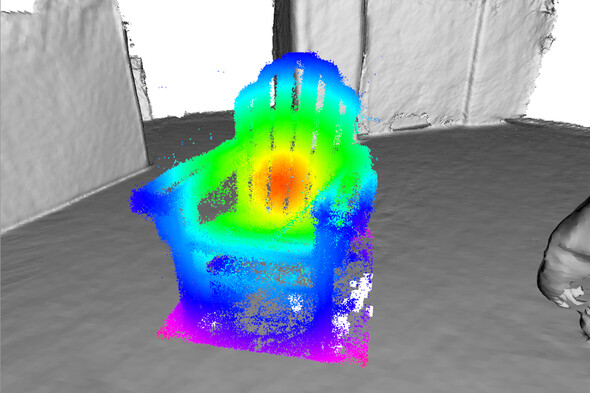}
        \hfill
        \includegraphics[width=0.19\columnwidth]{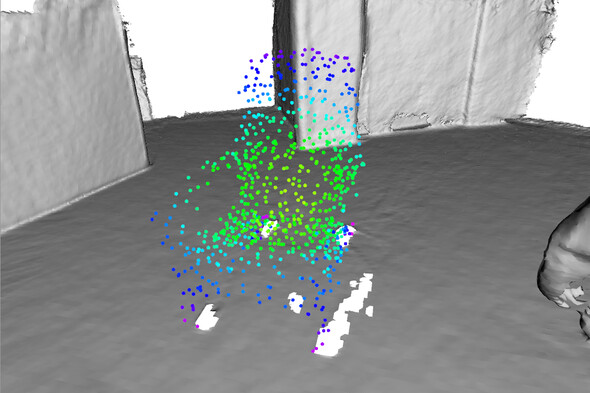}
        \hfill
        \includegraphics[width=0.19\columnwidth]{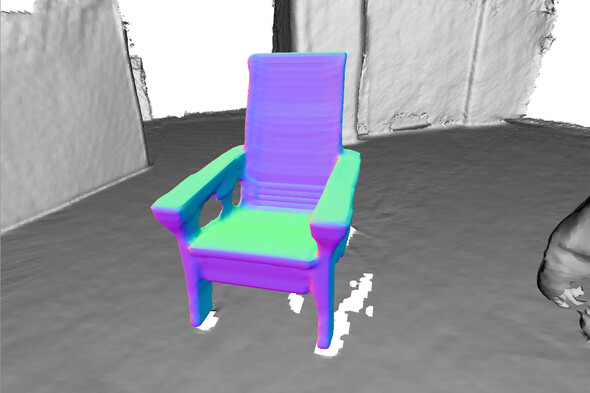}
    \end{subfigure}
    
    \begin{subfigure}[t]{\textwidth}
        \centering
        \includegraphics[width=0.19\columnwidth]{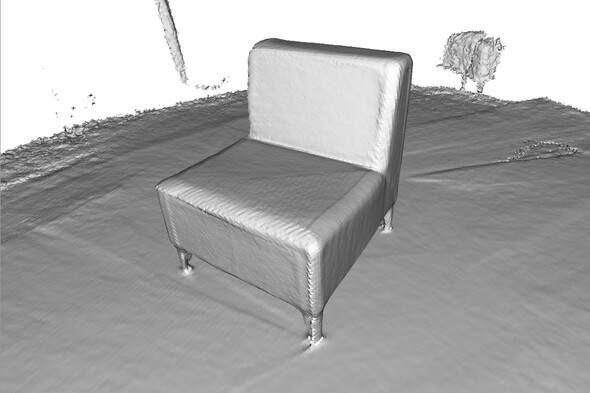}
        \hfill
        \includegraphics[width=0.19\columnwidth]{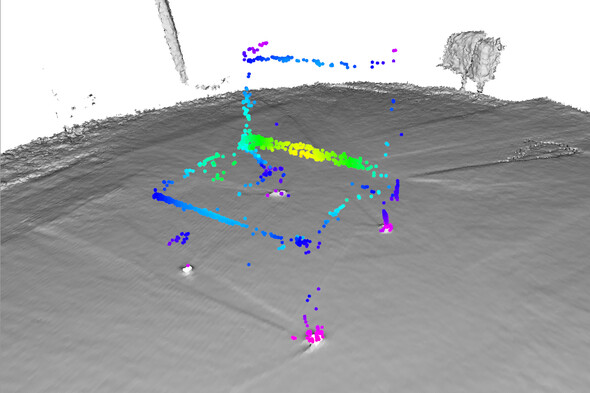}
        \hfill
        \includegraphics[width=0.19\columnwidth]{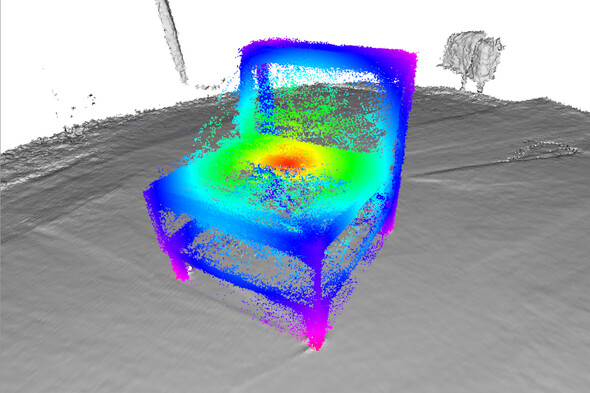}
        \hfill
        \includegraphics[width=0.19\columnwidth]{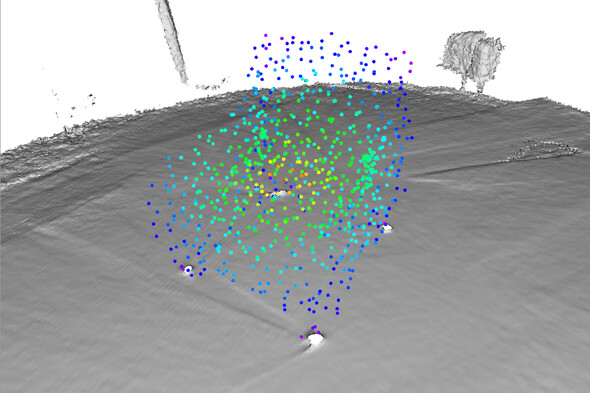}
        \hfill
        \includegraphics[width=0.19\columnwidth]{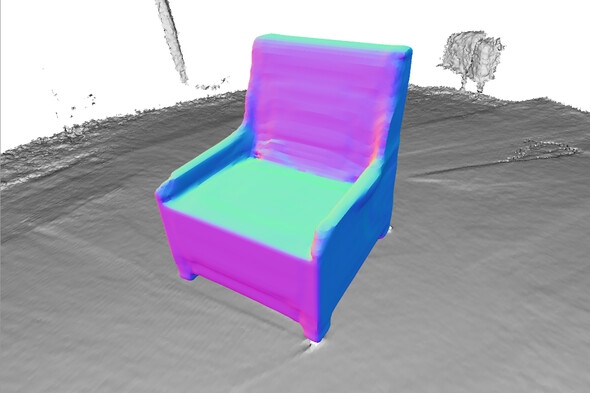}
    \end{subfigure}
    
    \begin{subfigure}[t]{\textwidth}
        \centering
        \includegraphics[width=0.19\columnwidth]{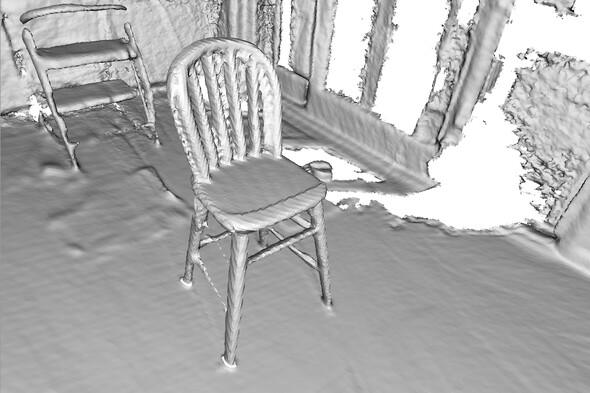}
        \hfill
        \includegraphics[width=0.19\columnwidth]{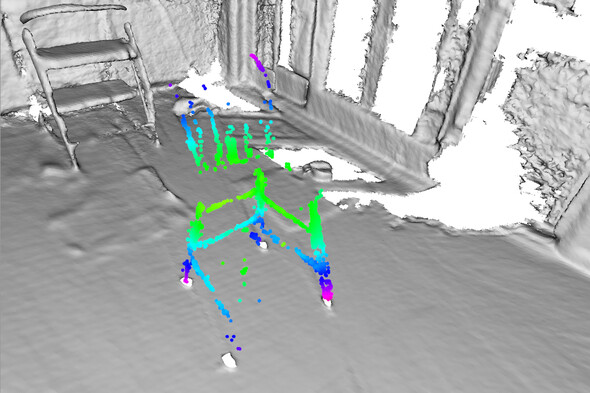}
        \hfill
        \includegraphics[width=0.19\columnwidth]{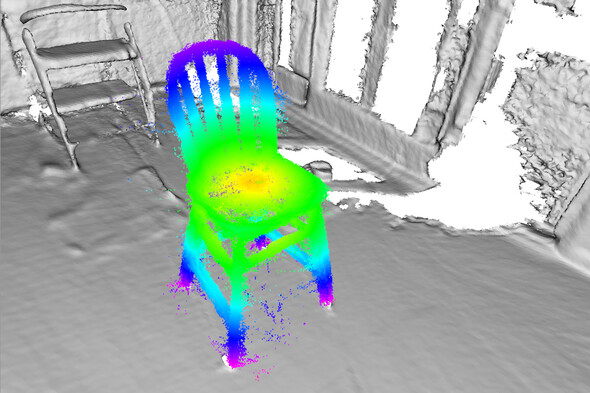}
        \hfill
        \includegraphics[width=0.19\columnwidth]{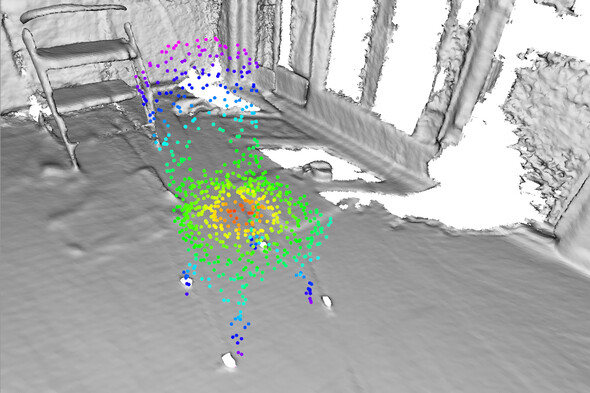}
        \hfill
        \includegraphics[width=0.19\columnwidth]{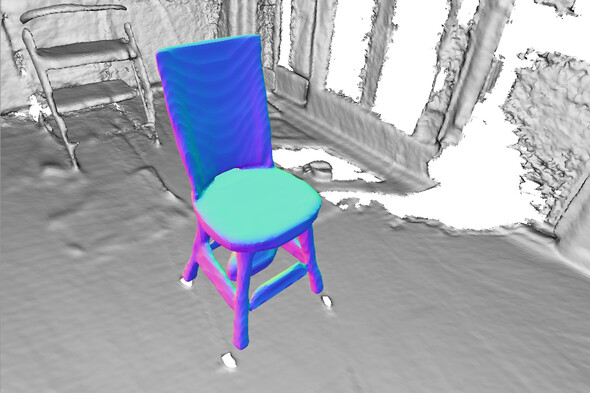}
    \end{subfigure}
    
    \begin{subfigure}[t]{\textwidth}
        \centering
        \includegraphics[width=0.19\columnwidth]{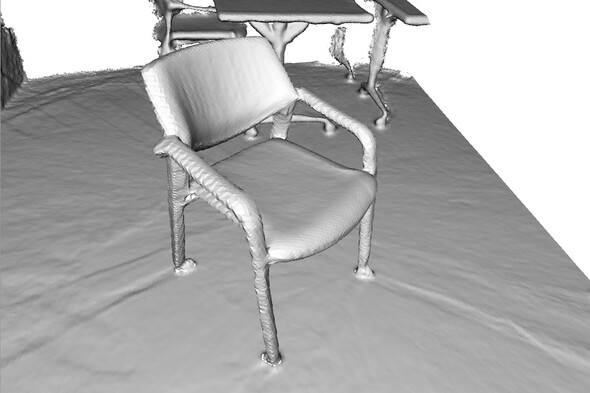}
        \hfill
        \includegraphics[width=0.19\columnwidth]{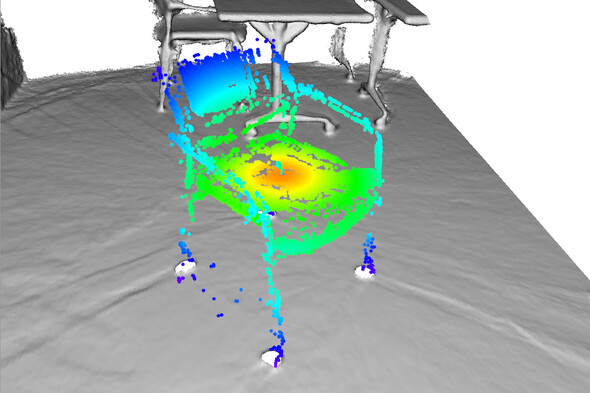}
        \hfill
        \includegraphics[width=0.19\columnwidth]{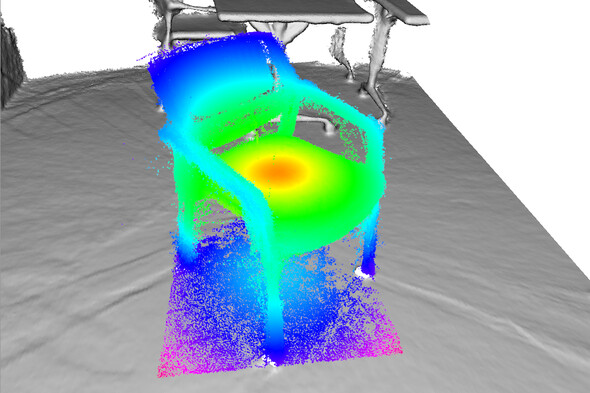}
        \hfill
        \includegraphics[width=0.19\columnwidth]{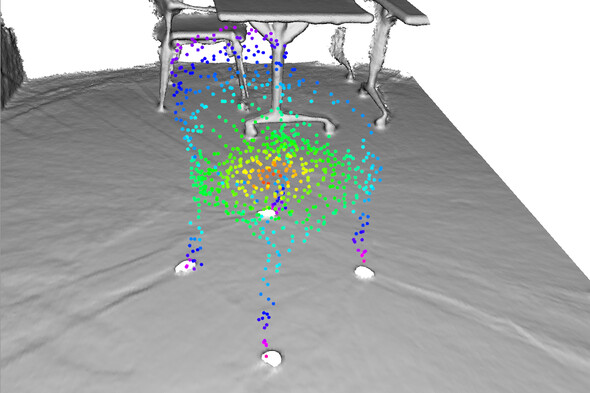}
        \hfill
        \includegraphics[width=0.19\columnwidth]{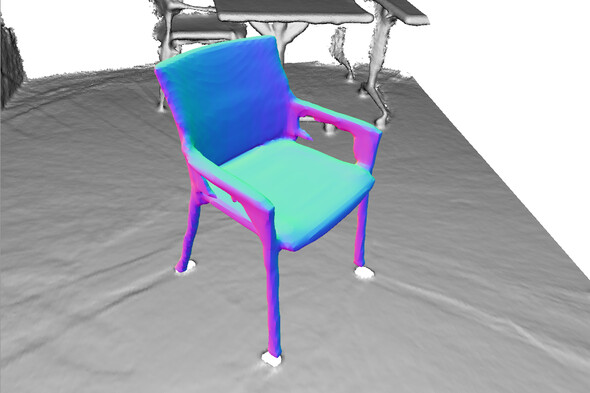}
    \end{subfigure}

    \begin{subfigure}[t]{\textwidth}
        \centering
\begin{tikzpicture}
      \tikzstyle{st}=[minimum width=34.8mm]
      \node [st] (11) {KinectFusion};
      \node[st,right = 0ex of 11] (12) {COLMAP (35)};
      \node[st,right = 0ex of 12] (13) {COLMAP (350)};
      \node[st,right = 0ex of 13] (14) {FroDO (sparse)};
      \node[st,right = 0ex of 14] (15) {FroDO (dense)};
\end{tikzpicture}
    \end{subfigure}
    
    \caption{Comparison of different approaches to object shape reconstruction
    on some examples from {\redwood} dataset. 
\label{fig:supp_redwood_result}}

\end{figure*}

Additional qualitative results on the \redwood{} dataset are shown in Figure~\ref{fig:supp_redwood_result}. As in the main paper, groundtruth, sparse COLMAP, dense COLMAP, our sparse reconstructions and our dense reconstructions are compared.

{\small
\bibliographystyle{ieee}
\bibliography{supp_materials}
}